\documentclass[12pt]{article}
\usepackage{amsmath}
\usepackage{amssymb}
\usepackage{graphicx}% Include figure files
\usepackage{bm}% bold math
\usepackage{multibib}
\usepackage{apacite}
\usepackage[margin=1.375in]{geometry}
\usepackage{authblk}

\begin{document}

%
%\usepackage{moderncvstyleclassic}
%\usepackage{hyperref}% add hypertext capabilities
%\usepackage[mathlines]{lineno}% Enable numbering of text and display math
%\linenumbers\relax % Commence numbering lines
%\usepackage[showframe,%Uncomment any one of the following lines to test 
%%scale=0.7, marginratio={1:1, 2:3}, ignoreall,% default settings
%%text={7in,10in},centering,
%%margin=1.5in,
%%total={6.5in,8.75in}, top=1.2in, left=0.9in, includefoot,
%%height=10in,a5paper,hmargin={3cm,0.8in},
%]{geometry}
\newlength{\figurewidth}
\setlength{\figurewidth}{0.7 \columnwidth}
\newcommand{\vsig}{{\vec{\sigma}}}
\newcommand{\la}{\left \langle}
\newcommand{\ra}{\right \rangle}

%\preprint{APS/123-QED}

\title{Knowledge as a Breaking of Ergodicity}

\author[$\dagger$]{Yang He}
\author[$\dagger$, $\star$, $\ddag$]{Vassiliy Lubchenko \thanks{vas@uh.edu}}
\affil[$\dagger$]{Department of Chemistry, University of Houston, Houston, TX 77204-5003}

%\author{Vassiliy Lubchenko}
%\email{vas@uh.edu}
%\affiliation{Department of Chemistry, University of Houston, Houston, TX 77204-5003}
\affil[$\star$]{Department of Physics, University of Houston, Houston, TX 77204-5005}
\affil[$\ddag$]{Texas Center for Superconductivity, University of Houston, Houston, TX 77204-5002}

\maketitle

\begin{abstract}

We construct a thermodynamic potential that can guide training of a generative model defined on a set of binary degrees of freedom. We argue that upon reduction in description, so as to make the generative model computationally-manageable, the potential develops multiple minima. This is mirrored by the emergence of multiple minima in the free energy proper of the generative model itself. The variety of training samples that employ $N$ binary degrees of freedom is ordinarily much lower than the size $2^N$ of the full phase space. The non-represented configurations, we argue, should be thought of as comprising a high-temperature phase separated by an extensive energy gap from the configurations composing the training set. Thus, training amounts to sampling a free energy surface in the form of a library of distinct bound states, each of which breaks ergodicity. The ergodicity breaking prevents escape into the near continuum of states comprising the high-temperature phase; thus it is necessary for proper functionality. It may however have the side effect of limiting access to patterns that were underrepresented in the training set. At the same time, the ergodicity breaking {\em within} the library complicates both learning and retrieval. As a remedy, one may concurrently employ multiple generative models---up to one model per free energy minimum.

\end{abstract}

\section{\label{sec:intro} Motivation}

Training sets and empirical data alike are often processed using representations that do not have an obvious physical meaning or are not optimized for the specific application computation-wise. Of particular interest are binary representations of information as would be pertinent to digital computation. Not only do the values of the binary variables depend on the detailed digitization recipe, but the number of training samples will usually be vastly smaller than the size $2^N$ of the full phase space available, in principle, to $N$ binary variables. Given this, one is justified in asking whether a {\em reduced} description exists that uses a relatively small number of variables and parameters to efficiently document the empirically relevant configurations. At the same time, it is desirable for the reduced description to be robust with respect to the choice of a discretization procedure that is used to present the original dataset. 

The problem of finding reduced descriptions is relevant for all fields of knowledge, of course, and is quite difficult in general. For example, the state of an equilibrated collection of particles is unambiguously specified by the expectation value of local density in a broad range of temperature and pressure~\cite{Evans1979}. Hereby particles exchange places on times comparable to or shorter than typical vibrational times, implying it is unnecessary to keep track of the myriad coordinates of individual particles. The equilibrium density profile is a unique, slowly varying function of just three spatial coordinates. Yet under certain conditions the translational symmetry becomes broken: One may no longer speak of an equilibrium density profile that is unique or smooth. Instead, one must keep track of a large collection of distinct, rapidly-varying density profiles each of which corresponds to a metastable solid; these profiles can be regarded as equilibrated with respect to particles' vibrations but not translations. For instance in a glassy melt~\cite{L_AP, LW_ARPC}, the number of alternative, metastable structures scales exponentially with the system size while the free energy surface becomes a vastly degenerate landscape that breaks ergodicity.

Here we address the problem of finding reduced descriptions in the context of machine learning. The {\em complete} description in the present setup is realized, by construction, through a generative model that is a universal approximator to an arbitrary digital dataset. That is, one can always choose such values for the model's parameters that the model will eventually have generated any given ensemble of $2^N$ distinct binary sequences of length $N$. The generative model is in the form of an Ising spin-based energy function, each spin representing a binary number. Ising spin-based generative models have been employed for decades~\cite{Hopfield1982, Laydevant2024, Mohseni2022}, of course. The present energy function has the functional form of the higher-order Boltzmann machine~\cite{sejnowski1986higher} and generally contains every possible combination of the spins. The learning rules are, however, different in that the coupling constants are deterministically expressed through the log-weights of individual sequences in the ensemble we want to reproduce. The retrieval is performed by Gibbs-sampling the Boltzmann distribution of the resulting energy function at a non-vanishing temperature. 

The present study begins by constructing effective thermodynamic potentials whose arguments are the parameters of the complete generative model. Each of these potentials is uniquely minimized by the {\em optimal} values of the coupling constants---in the complete description---and can be generalized so as to reflect correlations among distinct sub-ensembles. We note that effective thermodynamic potentials for a variety of generative models have been considered in the past. These are exemplified by the Helmholtz machine~\cite{DayanHintonNealEtAl95} or the loss function for the restricted Boltzmann machine~\cite{montufar2018restricted}, among others. 

Specific inquires during retrieval in the present generative model are made by imposing a constraint of user's choice. Of particular interest are constraints in the form of an additive contribution to the energy function that can stabilize a particular combination of the spins. This is analogous to how in a particle system, one can use an external---or ``source''---field to stabilize a desired density profile~\cite{L_AP, Evans1979}. If the system is ergodic, one may then use a Legendre transform to obtain a free energy as a function of the density profile. The latter procedure is a way to obtain a description in terms of variables of interest. At the same time, it represents a type of coarse-graining. Likewise, here we employ appropriate source-fields to produce a description in terms of variables of interest. The new degrees of freedom reflect weighted averages of the original spin degrees; thus their energetics are governed by a {\em free} energy. We show that when the source fields are turned off, this free energy becomes the aforementioned thermodynamic potential for the coupling constants. Thus learning and retrieval, respectively, can be thought of as minimizations on a conjoint free energy surface.

%The motivation behind considering such machines is their consistency over the long term: Any configuration of the machine, including sub-optimal ones, will be eventually escaped if the temperature is above zero.

The total number of the coupling constants in the complete description, $2^N$, becomes impractically large already for trivial applications, which then prompts one to ask whether the description can be reduced in some controlled way. The most direct way to reduce the description is to simply omit some terms from the energy function; the number of such terms increases combinatorially with the order of the interaction. We show that following a reduction in description, however, the free energy will increase non-uniformly, over the phase space, so as to develop multiple minima of comparable depth. By co-opting known results from Statistical Mechanics, we argue that, depending on the application, the amount of minima could be so large as to scale exponentially with the number of variables. 

The multiplicity of minima makes the choice of an optimal description ambiguous. Conversely, a successful reduction in description~\cite{Merchan2016, doi:10.1021/jp212541y} implies the underlying interactions are low-rank.  

Using the conjoint property of the free energy of learning and retrieval, respectively, we provide an explicit, rather general illustration of how the appearance of multiple minima in the learning potential also signals a breaking of ergodicity in the phase space spanned by the actual degrees of freedom of the model. Hereby the phase space of the spin system becomes fragmented into regions separated by free energy barriers \cite{Goldenfeld}. Consequently, escape rates from any given free energy minimum can become very low because they scale exponentially with the temperature and the parameters of the model~\cite{CL_LG}. Ergodicity breaking has been observed in restricted Boltzmann machines~\cite{10.21468/SciPostPhys.14.3.032, PhysRevLett.127.158303}, thus suggesting the latter machines represent reduced descriptions.  

The ergodicity breaking implies that quantities such as the energy and entropy, among others, are no longer state functions; instead, they could at best be thought of as multi-valued functions. Hence the notions of free energy and entropy---as well as the associated probability distribution---all become ill-defined. Indeed, having already two alternative free energy minima in a physical system corresponds to a coexistence of two distinct physical phases. Consider, for example, water near liquid-vapor coexistence at the standard pressure. The entropy of the system can vary by about $10 k_B$ per particle, depending on the phase, according to the venerable Trouton's rule for the entropy of boiling~\cite{BRR}. Thus the entropy and the free energy, as well as many other thermodynamic quantities are poorly defined. The uncertainty in the density of the system is particularly dramatic, viz. about three orders in magnitude \cite{LSurvey}.

As a potential remedy for the fragmentation of the space of the coupling constants caused by the reduction in description, one may employ a separate generative model for an individual minimum, or ``phase.'' Such individual models would each be ergodic and can be thought of as a Helmholtz machine~\cite{DayanHintonNealEtAl95}, since free energies can be defined unambiguously for a single-phase system. Running the machine would be, then, much like consulting multiple experts at the same time. Such multi-expert inquiries may in fact be unavoidable when distinct models employ sufficiently dissimilar interactions. Consider liquid-to-solid transitions as an example: Not only are such transitions intrinsically discontinuous~\cite{L_AP}, practical descriptions of the two respective phases involve altogether different variables~\cite{Lfutile}, as alluded to above. 

Since the number $2^N$ of possible configurations of $N$ bits becomes huge already for modest values of $N$, the vast majority of all possible configurations of the system are automatically missing from the training set. Because of their thermodynamically large entropy, we argue, these non-represented configurations can overwhelm the output of the machine. To avoid instabilities of this sort, the non-represented states must be placed at higher energies than the represented configurations; furthermore, the two sets of configurations should be separated by an extensive energy gap. Thus the missing configurations comprise a higher-temperature phase; this {\em also} implies a breaking of ergodicity. The robustness of the generative model, then, comes down to preventing a transition toward this high-temperature phase. At a fixed temperature, such stability is achieved by parameterizing the spectrum of the non-represented states so as to make the energy gap sufficiently large. Conversely at a fixed value of the gap, the retrieval temperature must be set below the transition temperature between the low-entropy and high-entropy phases, respectively. This will, however, limit one's ability to retrieve those configurations that were relatively underrepresented in the training set.

The article is organized as follows: In Section \ref{thermo} we construct a conjoint free energy surface whose arguments are the parameters of the generative model and, at the same time, coarse-grained values of the original degrees of freedom the complete description operates on. The formalism allows one to standardize weights of individual configurations in a dataset, as a means to implement calibration of sensors as well as manage bias or duplication in the data, if any. In Section~\ref{reduced}, we find ergodicity breaking already in the simplest possible realization of a reduced description. We make a connection with the ergodicity breaking during the physical phenomenon of phase coexistence and discuss implications for machine learning. Section~\ref{falsehood} provides a thermodynamically consistent treatment of incomplete datasets and argue that knowledge can be thought of as a library of bound states whose free energy is lower than the free energy of the non-represented states. Section~\ref{sec:conclusion} provides a summary and some perspective.

\section{\label{thermo} Thermodynamics of Learning and Retrieval}

\subsection{Setup of the generative model}

By construction, we consider a machine that operates on $N$ binary variables in the form of Ising spins $\sigma_\alpha = \pm 1$, $\alpha = 1, 2, \ldots, N$. The ``positive'' (``negative'') polarization state of each binary variable may be referred to as the ``up'' (``down'') state. These names are meant to specify the state of a binary register or a Boolean variable, as in the table below:
\begin{equation}
    \begin{tabular}{|c||c||c||c||c|}
    \hline
    $\sigma$ & polarization & spin-state & arrow & Boolean \\ \hline \hline
         $+1$ & positive & up & $\uparrow$ & 1 \\ \hline
         $-1$ & negative & down & $\downarrow$ & 0
         \\ \hline
    \end{tabular}
\end{equation}

The $2^N$ possible configurations of $N$ Ising spins $\sigma_\alpha$, $\alpha = 1, \ldots N$ cover the corners of a hyper-cube in an $N$-dimensional Hamming space. A configuration $i$ will be denoted as $\vec \sigma_i$:
\begin{equation} \label{sigidef}
     (\sigma_1^{(i)}, \sigma_2^{(i)}, \ldots, \sigma_N^{(i)}) \equiv \vsig_i \end{equation}
We will use Greek indexes to distinguish the $N$ directions in the Hamming space---these directions pertain to the individual bits (spins) themselves and correspond to the subscripts on the l.h.s. of Eq.~(\ref{sigidef}). 

We define a {\em dataset} by assigning a number $Z_i$ to configuration $i$ of the $N$ binary variables.  We further define the normalized weight of configuration $i$ according to
\begin{equation} \label{weights}
    x_i \equiv \frac{Z_i}{Z} % \equiv \frac{Z(\vsig_i)}{Z} 
\end{equation}
where 
\begin{equation}
    Z \equiv \sum_{i=1}^{2^N} Z_i, %\equiv \sum_{i=1}^{2^N} Z(\vsig_i) 
\end{equation}
the summation being over the $2^N$ points comprising the Hamming space. We will consistently label the latter points in the Hamming space using Latin indices, as well as any other quantities assigned to those points. 

By construction, the weight $x_i$  is intended to specify what a chemist would call the ``mole fraction'' of configuration $i$. The weight $x_i$ may or may not be associated with a probability, depending on the context; the weights are assigned using a user-defined convention as would be mole fractions in Chemistry, where one must explicitly specify what is meant by a ``species,'' ``particle'' etc. We set aside until Section~\ref{falsehood} the obvious issue that already for a very modestly-sized system, obtaining or storing $2^N$-worth of quantities $Z_i$ is impractical. For now, we simply assume that for those configurations not represented in the dataset, the respective weights are assigned some values of one's choice. In any event, the quantities $Z_i$ do not have to be integer. For instance, pretend we are teaching a machine, by example, the behavior of the inverter gate. Four distinct configurations are possible, in principle: (1) $\uparrow \downarrow$, (2) $\downarrow \uparrow$, (3) $\uparrow \uparrow$, and (4) $\downarrow \downarrow$, where one arrow stands for the input bit and the other arrow for the output bit. For concreteness, let us set $Z_i$ at the number of times configuration $i$ was presented in the set. Suppose the training set is $Z_1 = 32$, $Z_2 = 37$, $Z_3 = 2$, $Z_4 = 0$. It will be useful to regard $Z_4$ as an adjustable parameter, even if we eventually adopt for it a fixed value that is very small relative to the rest of the $Z_i$'s. 

We will sometimes refer to expressions of the type
\begin{equation} \overline f \equiv \sum_i x_i f_i
\end{equation}
as ``weighted sums,'' or ``averages,'' or ``expectation values.''  

We define a {\em generative model} in the form of an energy-like function acting on $N$ binary degrees of freedom. Because $\sigma^{2n +1} = \sigma$ for integer $n$, the most general function of the spin variables $\sigma_\alpha$ can be written as a linear combination of all possible products of the variables, where in each product, a particular component $\sigma_\alpha$ is present at most once. Thus one may define the following function
\begin{eqnarray} \label{EGM}
    E(\vsig) & \equiv &- J_0 - \sum_{\alpha_1}^N J_{\alpha_1} \, \sigma_{\alpha_1} - \hspace{-1mm} \sum_{\alpha_1 < \alpha_2}^N J_{\alpha_1 \alpha_2} \, \sigma_{\alpha_1} \sigma_{\alpha_2} \nonumber - \hspace{-2mm}\sum_{\alpha_1 < \alpha_2 < \alpha_3}^N J_{\alpha_1 \alpha_2 \alpha_3} \, \sigma_{\alpha_1} \sigma_{\alpha_2} \sigma_{\alpha_3}  - \ldots \nonumber \\ &\equiv& - J_0 - \sum_{n=1}^{N} \: \sum_{\alpha_1 < \ldots < \alpha_n}^N J_{\alpha_1 \alpha_2 \ldots \alpha_n} \prod_{i=1}^n \sigma_{\alpha_i} \nonumber \\ &\equiv& - \langle \bm J | \bm \sigma \rangle
\end{eqnarray}
A parameter $J_{\alpha_1 \alpha_2 \ldots \alpha_n}$ is the coupling constant for the interaction that couples the $n$ spins
$\sigma_{\alpha_1}$, $\sigma_{\alpha_2}$, $\ldots$,
$\sigma_{\alpha_n}$. The last line in Eq.~(\ref{EGM}) purveys a useful
shorthand whereby we present the energy function as the inner product
of two $2^N$-dimensional vectors ${\bm J}$ and ${\bm \sigma}$. By
construction, vector ${\bm \sigma}$ is the Kronecker product of all
pairs $(1, \sigma_\alpha)$:
\begin{equation} \label{Kron}
    {\bm \sigma} \equiv (1, \sigma_N) \otimes (1, \sigma_{N-1}) \otimes \ldots \otimes (1, \sigma_1). 
\end{equation}
We will consistently order all Kronecker products so that the spin label $\alpha$ increases right to left. For instance, for three spins one has ${\bm \sigma} = (1, \sigma_3) \otimes (1, \sigma_2) \otimes (1, \sigma_1) =  (1, \, \sigma_1, \,\sigma_2, \, \sigma_1 \sigma_2, \, \sigma_3, $ $ \sigma_1 \sigma_3, \, \sigma_2 \sigma_3, \, \sigma_1 \sigma_2 \sigma_3)$.  The components of the $2^N$-dimensional vector ${\bm J}$ are labeled so as to match the combination of spins the component in question multiplies, per Eq.~(\ref{EGM}). ($J_0$ multiplies the $1$.) And so for three spins one has ${\bm J} \equiv (J_0, \, J_1, \, J_2, \, J_{12}, \, J_3, \, J_{13}, \, J_{23}, \, J_{123})$ etc. 

Owing to the mixed-product property of the Kronecker product, two vectors $\bm \sigma_i$ and $\bm \sigma_j$ are orthogonal, if $\sigma_\alpha^{(i)} \ne \sigma_\alpha^{(j)}$ at least for one $\alpha$. Thus,
\begin{equation} \label{ortho}
    \la {\bm \sigma}_i | {\bm \sigma}_j \ra = 2^N \delta_{ij}.
\end{equation}
%which can also be viewed as a consequence of the string $(1, \sigma) = \sigma (\sigma, 1)$ being an irreducible representation of the symmetric group $S_2$, on the one hand, and Eq.~(\ref{Kron}), on the other hand. 
This orthogonality relation, in turn, implies the following completeness relation:
\begin{equation} \label{completeness}
    \sum_i | \bm \sigma_i \rangle \langle \bm \sigma_i | = 2^N \: {\bm 1},
\end{equation}
where $|\bm  a \rangle \langle \bm b |$ denotes the outer product of vectors $\bm a$ and $\bm b$ and ${\bm 1}$ is the unit matrix of size $2^N$.

Because the total number $\sum_{n=0}^N N!/n! (N-n)! = 2^N$ of the coupling constants matches the total number of configurations available to our spin system, one may inquire if there is a set of coupling constants such that the set of the $2^N$ values of the function $E(\vec \sigma_i)$, $i=1, \ldots, 2^N$, can match exactly an  arbitrary set of energies $E_i$, $i=1, \ldots, 2^N$: $E(\vec \sigma_i) = E_i$. Eq.~(\ref{EGM}), then, implies the coupling constants would have to solve the following system of $2^N$ linear equations:
\begin{equation}
    \la {\bm J} | {\bm \sigma}_i \ra = - E_i \hspace{3mm}  (i = 1, \ldots 2^N).
\end{equation}
The solution indeed always exists, is unique, and
is straightforwardly obtained by multiplying the above equation by $\langle \bm \sigma_i|$ on the right, summing over $i$, and using Eq.~(\ref{completeness}), see also \cite{gresele2017maximum}.  This yields
\begin{equation} \label{JlnZ}
    \bm J 
    %= \frac{T}{2^N} \sum_{i=1}^{2^N} \bm \sigma_i \ln Z_i 
    = - \frac{1}{2^N} \sum_{i=1}^{2^N} \bm \sigma_i E_i 
\end{equation}
or, more explicitly,
\begin{equation} \tag{\ref{JlnZ}$'$}\label{JlnZprime}
\begin{split}
    J_0 &
    = - \frac{1}{2^N} \sum_{i} E_i \\
    J_\alpha &
    = - \frac{1}{2^N} \sum_{i} \sigma_{\alpha}^{(i)} E_i \\
    J_{\alpha \beta} &
    =-  \frac{1}{2^N} \sum_{i} \sigma_{\alpha}^{(i)}  \sigma_{\beta}^{(i)} E_i \\
    J_{\alpha \beta \gamma} &
    = - \frac{1}{2^N} \sum_{i} \sigma_{\alpha}^{(i)}  \sigma_{\beta}^{(i)} \sigma_{\gamma}^{(i)} E_i  \\
    \cdots & 
\end{split}
\end{equation}

%\subsection{Connecting the generative model with the dataset}

It is not immediately obvious how much  the coefficients $\bm J$ from Eq.~(\ref{JlnZ})---and hence the generative model itself---would vary among datasets originating from distinct sources/experiments that one may, nonetheless, deem being qualitatively similar or even equivalent. In the happy event that the variation is indeed small, such robustness could be thought of, by analogy with thermodynamics, as the couplings $\bm J$ being subject to a smoothly varying, free-energy surface. With the aim of constructing such a free-energy surface, we next discuss a variety of ways to connect our generative model with a dataset and, conversely, how to retrieve patterns learned by the model.  %In addition, the weights $x_i$ may be either discrete or continuous variables. If the weights are in fact discrete variables, connecting them in a smooth fashion with the coupling constants ${\bm J}$ requires a course-graining recipe; the aforementioned free energy surface will serve to this end as well. 

\subsection{Data retrieval and calibration. Learning rules}

 We consider a candidate ensemble in which the numbers $Z_i$ are driven by source fields in the form of the energies $E_i$ themselves. The couplings $\bm J$---which are ultimately of interest---can be then determined using the relations (\ref{JlnZ}). We are specifically interested in the ability to drive the distribution $Z_i$ with respect to the energy $E_i$ relative to some preset reference value $E_i^\ominus$:
\begin{equation} \label{ZE}
    Z_i = Z_i^\ominus e^{-(E_i-E_i^\ominus)/T} = e^{-[E_i-(E_i^\ominus-T \ln Z_i^\ominus)]/T},
\end{equation}
We will call the reference values ``standard'' and consistently label them with the symbol $\ominus$. By construction, $Z_i^\ominus$ and $E_i^\ominus$ are independent of temperature.

Retrieval in the setup above is performed by Gibbs sampling, at temperature $T$, the energy function obtained by computing $E_i = E(\vec \sigma_i)$ with the help of  (\ref{EGM}), and then subtracting from it the quantity $(E_i^\ominus-T \ln Z_i^\ominus)$. If parallel tempering is used, the quantity $T$ in the last expression is fixed at the target temperature, of course.

The standard value $Z_i^\ominus$ serves to calibrate the weight of configuration $i$, if presented, in the dataset; the quantities $Z_i^\ominus$ are non-vanishing, by construction. Let us provide several examples of where such calibration is useful or even necessary: (a)~If distinct configurations are detected using separate sensors, the outputs of the latter sensors must be calibrated against each other. To drive this point home, imagine that two or more detectors operate using different physical phenomena. For instance, one may determine temperature by using a known equation of state for a material or by spectroscopic means, among many others. Outputs of distinct detectors can be and must be mutually calibrated where the respective ranges of detection overlap. (b)~Calibration may be necessary to manage duplication in the dataset or a bias, if any, in the acquisition or publication of the data; see \cite{doi:10.1021/jacs.1c12005} for a discussion of such a bias in the context of using machine learning to predict optimal reaction conditions. (c)~One can think of the quantity $\ln Z_i^\ominus$ as the intrinsic entropy of state $i$, according to the second equality in Eq.~(\ref{ZE}). The latter entropy could, for instance, reflect the log-number of states of a hidden degree of freedom, when the {\em visible} variables happen to be in configuration $i$. (d)~Variation with respect to the ``local'' energy difference $(E_i-E_i^\ominus)$ is analogous to the material derivative, whereby the local reference energy $E_i^\ominus$ can be thought of specifying context. This is useful if the inputs exhibit correlations and/or one wishes to parameterize differences among distinct inputs, see Appendix~\ref{correlations}. There we also compare the present use of standard values to that in Chemistry. Last but not least, one may choose to regard the standard values $Z_i^\ominus$ and $E_i^\ominus$ as inherently distributed, in a Bayesian spirit.

According to Eq.~(\ref{ZE}), the ``local'' energy deviations $(E_i-E_i^\ominus)$ reflect how much the coupling constants $\bm J$ would have to be perturbed from their standard values to shift the distribution $\{ Z_i \}$ from its standard value. Suppose, for the sake of argument, the latter shift is due to incoming additional data and that we set $Z_i$ at the number of instances of configuration $i$. Then Eqs.~(\ref{JlnZ}) effectively prescribe a set of learning rules. For instance, a two-body coupling constant will be modified according to $J_{\alpha \beta} \to J_{\alpha \beta} - (1/2^N) \sum_{i} \sigma_{\alpha}^{(i)} \sigma_{\beta}^{(i)} (E_i - 
E_i^\ominus) = J_{\alpha \beta} + (T/2^N) \sum_{i} \sigma_{\alpha}^{(i)} \sigma_{\beta}^{(i)}  \ln (Z_i/Z_i^\ominus)$. We see this learning rule is similar to, but distinct from the venerable Hebbian rule, in which instances would, instead, add up cumulatively. Here, in contrast, the coupling constants are weighted by the log-number of instances of reinforcement. 

\subsection{Free energy of learning}

Consider the following object:
\begin{align} \label{Abasic}
    A(\{ E_i \}, T) \equiv & - T \ln \left( \sum_i Z_i \right) = - T \ln  \left( \sum_i Z_i^\ominus e^{-(E_i-E_i^\ominus)/T} \right),
\end{align}
where we treat each quantity $Z_i = Z_i^\ominus e^{-(E_i-E_i^\ominus)/T}$ and the quantity $A$ itself as functions of the energies $E_i$ and temperature $T$. We next consider the following quantity:
\begin{align}
    \widetilde S \equiv - \left( \frac{\partial A}{\partial T}\right)_{E_i} =&  \ln Z - \sum_i \left( - \frac{E_i-E_i^\ominus}{T} \right) \frac{Z_i}{Z} \nonumber  = \ln Z^\ominus - \sum_i \left(\ln \frac{Z_i/Z}{Z_i^\ominus/Z^\ominus} \right) \frac{Z_i}{Z}  \\ =& \widetilde S^\ominus - \sum_i \left(\ln \frac{x_i}{x_i^\ominus} \right) x_i \label{S1} \\ \le & \widetilde S^\ominus \label{Sineq}
\end{align}
In Eq.~(\ref{S1}), we reflected that the sum vanishes for the standard weights and used the definition  (\ref{weights}) of the weights $x_i$. The inequality in Eq.~(\ref{Sineq}) holds because the sum in the above equation is non-negative. (The latter sum also happens to be the Kullback–Leibler divergence between the distributions defined by $x_i$ and $x_i^\ominus$, respectively.) The divergence is minimized---and hence the quantity $\widetilde S$ is maximized---along the line $Z_i/Z = Z_i^\ominus/Z^\ominus \equiv x_i^\ominus$. If one imposes an additional constraint, for instance by fixing the energy-like quantity
\begin{equation} \label{E}
    \widetilde E \equiv A + T \widetilde S  = \sum_i x_i \left(E_i-E_i^\ominus \right),
\end{equation}
the quantity $\widetilde S$ is now maximized by a {\em unique} configuration of $\{Z_i\}$ or, equivalently, of the energies $\{ E_i \}$. Thus the quantity $\widetilde S$ is an {\em entropy}. 

Clearly $\widetilde E^\ominus = 0$; consequently
\begin{equation} \label{Astandard}
    A^\ominus \equiv A(\{ E_i^\ominus \}, T) = - T \widetilde S^\ominus.
\end{equation}
This implies the standard entropy $\widetilde S^\ominus$ is temperature independent, since according to Eq.~(\ref{Abasic}), the standard value $A^\ominus$
\begin{equation}
  A^\ominus = - T \ln \sum_i Z_i^\ominus \equiv - T \ln Z^\ominus
\end{equation}
is strictly proportional to temperature. Also, Eqs.~(\ref{Astandard}) and (\ref{Abasic}) can be used to show that in the  gauge $Z_i^\ominus = e^{-E_i^\ominus/T^\circ}$, where $T^\circ$ is a positive constant, the standard entropy is given by the following expression:   
\begin{equation} \label{shannon}
    \widetilde S^\ominus = - \sum_i x_i^\ominus \ln x_i^\ominus - \sum_i x_i^\ominus E_i^\ominus/T^\circ, \hspace{3mm} \text{if } Z_i^\ominus = e^{-E_i^\ominus/T^\circ}.
\end{equation} 
The entropy above has the form of the Shannon entropy of the dataset, if one sets $\sum_i x_i^\ominus E_i^\ominus  = 0$. The latter can be done without loss of generality but will have the effect of fixing the energy reference.

One may also readily define a heat capacity-like quantity:
\begin{equation}
    \widetilde C \equiv \left( \frac{\partial \widetilde E}{\partial T}\right)_{E_i} \!\! = T \left( \frac{\partial \widetilde S}{\partial T}\right)_{E_i} \!\! = \sum_i x_i \frac{\left(E_i-E_i^\ominus \right)^2 - \widetilde E^2}{T^2} 
\end{equation}
The quantities $\widetilde E$ and $\widetilde C$ can be regarded as analogs of the conventional energy and heat capacity, respectively, but generalized for a local ``gauge'' in the form of the state-specific energy references $E_i^\ominus$. Conversely, if we adopt a global gauge, by adopting a uniform $E_i^\ominus=\text{const}$, we obtain the conventional forms $\widetilde E = \overline E - \text{const}$ and $\widetilde C = (\overline{E^2} - \overline E^2)/T^2$, respectively. 

According to Eqs.~(\ref{Abasic}) and (\ref{weights}):
\begin{equation} \label{AEx}
    \left. \frac{\partial A}{\partial E_i} \right|_{E_{j\ne i, T}} = \frac{Z_i}{Z} = x_i.
\end{equation}
Thus we obtain that the function $A$ is a free energy:
\begin{equation}
    dA = - \widetilde S dT + \sum_i^{2^N} x_i \, dE_i.
\end{equation}
Consistent with this identification, $A(\{ E_i \}, T=\text{const})$ is a convex-up function of the variables $\{ E_i \}$. The curvature vanishes in exactly one direction, $Z_i/Z = \text{const}$; movement in the latter direction leaves the weights $x_i$ invariant. %$A(\{ E_i \}, T=\text{const})$ does become strictly convex-up in the presence of a constraint, such as the aforementioned constraint $\widetilde E = \text{const}$.

A more useful description is afforded by a free energy that operates on the weights $x_i$---which are the actual ``observable'' quantities; we accomplish this using Legendre transforms. Choose any number $M$ of configurations of interest, and assign to them  labels $1$ through $M$. One may then define the following Legendre transform, whereby one removes the energetic contribution to the shift of the free energy off its standard value, while retaining exclusively the entropic contribution, for configurations $1$ through $M$:
%\begin{equation}
%    \Phi^{(M)}(\{ x_i \}, T) \equiv A - \sum_i^M E_i x_i 
%\end{equation}
%leading to
%\begin{equation} \label{dF}
%    d\Phi^{(M)} = - \widetilde S^{(M)} dT - \sum_i^M E_i \, d x_i. 
%\end{equation}
\begin{equation}
\label{FtA}
    \widetilde A^{(M)}(\{ x_i \}, T) \equiv  A - \sum_i^M (E_i - E_i^\ominus) x_i, 
\end{equation}
so that 
\begin{equation} \label{dAt}
    d \widetilde A^{(M)} = - \widetilde S^{(M)} dT - \sum_i^M (E_i - E_i^\ominus) \, d x_i. 
\end{equation}
Thus,
\begin{equation} \label{Ex}
    E_i = E_i^\ominus -\left( \frac{\partial \widetilde A^{(M)}}{\partial x_i} \right)_{x_j, E_k, T} \hspace{5mm} (i, j \le M, j \ne i; k > M)
\end{equation}
and we have defined
\begin{equation} \label{Sx}
    \widetilde S^{(M)} \equiv  -\left( \frac{\partial \widetilde A^{(M)}}{\partial T} \right)_{x_i, E_k} \hspace{5mm} (i \le M, k > M).
\end{equation}
We note that
\begin{equation}
\widetilde S^{(M)} = \widetilde S \hspace{5mm} \text{when } M=2^N.
\end{equation}
Also, $\sum_i^M Z_i = Z \sum_i^M x_i$; thus $Z = Z'/(1 - \sum_i^M x_i)$ and, subsequently,
\begin{equation} \label{Zixi}
    Z_i = Z' \frac{x_i}{1 - \sum_i^M x_i},
\end{equation}
which can be used to express $\widetilde A^{(M)}$ explicitly in terms of $x_i$: 
\begin{equation}
\begin{split}
\label{Atilde}
    \widetilde A^{(M)}  = 
    \left\{ \begin{array}{l} T \left[\sum_i^M x_i \ln\frac{x_i}{x_i^\ominus} + \left(1 - \sum_i^M x_i \right) \ln \frac{1-\sum_i^M x_i}{x'^\ominus}\right]  - T \ln Z^\ominus,  \hspace{5mm} (M < 2^N) \\ \\
    T \sum_i x_i \ln\frac{x_i}{x_i^\ominus} - T \ln Z^\ominus,  \hspace{5mm} (M = 2^N) \end{array} \right.
\end{split}
\end{equation}
where 
\begin{equation}
    x' \equiv Z'/Z.
\end{equation}
and $\sum_i x_i =1$, of course. Throughout, summation over configurations spans the range $1$ through either the explicitly indicated upper limit or, otherwise, through $2^N$. We will use the following shorthand below:
\begin{equation}
    \widetilde A \equiv \widetilde A^{(M)} \hspace{5mm} \text{when } M=2^N.
\end{equation}
Further, Eqs.~(\ref{Sx}) and (\ref{Atilde}) imply
\begin{equation} \label{ATS}
    \widetilde A^{(M)} = - T \widetilde S^{(M)}.
\end{equation}
Consequently, $d \widetilde A^{(M)} = - T d \widetilde S^{(M)} - \widetilde S^{(M)} dT$. Combining this with Eq.~(\ref{dAt}) yields
\begin{equation}
    d \widetilde S^{(M)} = \sum_i^M \frac{E_i-E_i^\ominus}{T} \, d x_i. 
\end{equation}
Thus the entropy $\widetilde S^{(M)}$ is uniquely maximized at $E_i = E^\ominus_i$  and, hence at $x_i = x_i^\ominus$. Consequently, the function $\widetilde A^{(M)}$ is uniquely minimized at $x_i = x_i^\ominus$, by Eq.~(\ref{ATS}):
\begin{equation} \label{Aineq}
    \widetilde A^{(M)} \ge A^\ominus = - T \ln Z^\ominus,
\end{equation}
for any $M$. These notions can be also established directly by differentiating Eq.~(\ref{Atilde}) with respect to the $x_i$'s. Thus the function $\widetilde A$ is analogous to a thermodynamic potential governing the relaxation of a single-phase physical system. Note Eqs.~(\ref{Atilde}) and (\ref{Aineq}) are consistent with Eq.~(2.6) of \cite{DayanHintonNealEtAl95}.

The bound in Eq.~(\ref{Aineq}) is connected to the familiar Gibbs inequality. Indeed, in view of Eq.~(\ref{FtA}), inequality (\ref{Aineq}) becomes
\begin{equation} \label{GibbsGen}
    A^\ominus \le A + \sum_i^M x_i (E_i^\ominus - E_i),
\end{equation}
At $M=2^N$, the above equation is the traditional way to express the Gibbs inequality~\cite{GirardeauMazoVariational}:
\begin{equation} \label{Gibbs}
    A^\ominus \le A + \sum_i x_i (E_i^\ominus - E_i).
\end{equation}
In other words, the free energy of a system of interest---represented by the standard model in this case---is bounded from above by the free energy for a trial energy function plus the expectation value of the correct energy, relative to the trial energy, where the weights are computed using the {\em trial} energy function. 

According to Eqs.~(\ref{JlnZ}) and (\ref{ZE}), there is one-to-one correspondence between the quantities $Z_i$ and the coupling constants $\bm J$. Thus Eqs.~(\ref{JlnZ}) can be viewed as the solution of a  minimization problem in which one minimizes the thermodynamic potential $\widetilde A$ with respect to the coupling constants $\bm J$, while keeping exactly one coupling constant fixed. (It is most convenient to fix $J_0$, which simply specifies an overall multiplicative factor for the numbers $Z_i$ and, hence, does not affect the weights $x_i$.). In this sense, one can think of $\widetilde A$ as being able to guide a learning process, in principle. %There are an infinite variety of potentials that are minimized by the same set of the coupling constants, one such example provided in the Appendix. Still, the specific potential from Eq.~(\ref{Atilde}) has the distinction of being a free energy as a function of appropriate coarse-grained variables; this will be of use in Section~\ref{reduced}.

What is the meaning of the distribution $e^{-\widetilde A(\{ x_i \})/T}$, which one may nominally associate with the potential $\widetilde A$? Analogous distributions arise in Thermodynamics as a result of the canonical construction. Hereby one effectively considers an infinite ensemble of distinct but physically equivalent replicas of the system \cite{McQuarrie}. The width of the distribution $e^{-\widetilde A/T}$ then should be formally thought of as variations of the weights $x_i$ among distinct replicas. %At the same time, the latter width is proportional to the susceptibility $(\partial^2 \widetilde A/\partial x_i \partial x_j)^{-1}$. Thus a large---let alone diverging---susceptibility would signal a breakdown of the canonical description. In statistical terms, this would mean the data do not exhibit a discernible trend. 
To generate such ensembles in practice one can, for instance, break up a very large dataset into smaller---but still large---partial datasets. In some cases data may exhibit correlations due to implicit or hidden variables, such as the time and place of collection. If so, breaking datasets into pertinent subsets may reveal correlations among {\em fluctuations} of the weights $x_i$---from subset to subset---that are not necessarily captured by the ideal mixture-like expression (\ref{Atilde}). Adopting a particular calibration scheme for the inputs of the sensors may {\em also} introduce correlations among weight fluctuations. We outline, in Appendix~\ref{correlations}, a possible way to modify the free energy so as to account for the latter correlations.

\subsection{Coarse-graining, choice of description, and inquiries}

Because spins are intrinsically discrete degrees of freedom, equivalent yet distinct datasets may result in spin arrangements that are similar in some course-grained sense, yet may be rather dissimilar at the level of individual spins owing, for instance, to noise. It may, then, be advantageous to specify the state of the system at a coarse-grained level, i.e., using not the discrete variables $\sigma_\alpha$---themselves or combinations thereof---but, instead, their averaged values that retain essential features of the pattern while not being overly sensitive to detailed variations in the polarization pattern. Here we use the coarse-graining recipe underlying the canonical construction whereby one applies a source field $\bm h$ to the spins, which yields the following, equilibrium Gibbs energy:
\begin{equation} \label{G}
    G =  - T \ln \left(\sum_i Z_i e^{\la \bm h | \bm \sigma_i \ra/T} \right).
\end{equation}
The source field $\bm h$ itself is independent of spin configuration and can be used to stabilize spin arrangements of interest. Throughout, we will limit ourselves to source fields in the form of a sum of one-spin, onsite fields:
\begin{equation} \label{hs}
    \la \bm h | \bm \sigma \ra = \sum_\alpha h_\alpha \sigma_\alpha.
\end{equation}
%Such a form could be used, for instance, to implement a pixelation scheme by uniformly applying the same source field to a compact set of pixels, that together will have comprised some new, larger pixels. %Other functional forms including products $\sigma_\alpha \sigma_\beta$, $\sigma_\alpha \sigma_\beta \sigma_\gamma$ etc., can be adopted as appropriate, so as to probe for multi-spin correlations. 

In a straightforward manner, one can determine the typical magnetization of spin $\alpha$ by varying the free energy $G$ with respect to the source field $h_\alpha$:
\begin{equation} \label{dGdh}
    - \frac{\partial G}{\partial h_\alpha}  = \sum_i x_i \sigma_\alpha^{(i)} \equiv \overline{\sigma}_\alpha = m_\alpha .
\end{equation}
where the weights $x_i$ now reflect the additional bias due to the external field: 
\begin{equation} \label{weightsh}
    x_i \equiv \frac{Z_i e^{\la \bm h | \bm \sigma_i \ra/T} }{\sum_j Z_j e^{\la \bm h | \bm \sigma_j \ra/T}  },  
\end{equation}
c.f. Eq.~(\ref{weights}). 

The simple type of source field on the r.h.s. of Eq.~(\ref{hs}) is just one particular---albeit instructive---functional form for a source field. An arbitrary linear combination of spin products $\sigma_\alpha$, $\sigma_\alpha \sigma_\beta$, $\sigma_\alpha \sigma_\beta \sigma_\gamma$, etc., can be stabilized by the same token. From a formal viewpoint, such complicated forms can be used to probe for multi-spin correlations. More significant in the present context, however, is that the corresponding expectation values constitute a description in terms of some new variables of choice. These variables are coarse-grained combinations of the original spin degrees of freedom, as already mentioned. Thus the appropriate energy quantity that determines the statistics of these coarse-grained variables is a {\em free} energy. Indeed, according to Eq.~(\ref{dGdh}), one may introduce the Helmholtz energy in the usual way:
\begin{equation} \label{FGm}
   A = G + \sum_\alpha h_\alpha m_\alpha.
\end{equation}
where the source fields $h_\alpha$ can be thought of as external fields that would be necessary to induce a magnetization pattern of interest:
\begin{equation} \label{hFM}
    h_\alpha = \frac{\partial A}{\partial m_\alpha}.
\end{equation}
At the same time, 
\begin{equation} \label{dGdE}
    \frac{\partial G}{\partial E_i}  = x_i, 
\end{equation}
c.f. Eq.~(\ref{AEx}), and so the quantity 
\begin{equation} \label{GtG}
    \widetilde G \equiv G - \sum_i x_i (E_i - E_i^\ominus) 
\end{equation}
should be recognized as the Gibbs counterpart of $\widetilde A$. 
%in the presence of non-vanishing source fields. with the source fields being the control variables:
%\begin{equation} \label{dGt}
%    d \widetilde G = - \widetilde S dT - \sum_i (E_i %- E_i^\ominus) \, d x_i. 
%\end{equation}

We make a connection with the thermodynamic potential $\widetilde A$ from Eq.~(\ref{Atilde}) by repeating the steps that led to Eq.~(\ref{Atilde}) while having replaced $E_i$ by the quantity $E_i^{(h)}$:
\begin{equation} \label{Eh}
    E_i^{(h)} \equiv E_i - \la \bm h | \bm \sigma_i \ra
\end{equation}
in Eq.~(\ref{FtA}) and $A$ by $G$. One thus obtains
\begin{equation} \label{GtAh}
    G - \sum_i \left(E_i^{(h)} - E_i^{(h)\ominus} \right) x_i = \widetilde A(\{ x_i \}).
\end{equation}
At the same time, Eqs.~(\ref{hs}), (\ref{dGdh}), and (\ref{Eh}) imply that
\begin{equation} \label{Ehav}
    \sum_i x_i E_i^{(h)}  = \sum_i x_i E_i - \sum_\alpha h_\alpha m_\alpha.
\end{equation}
Combined with Eq.~(\ref{GtAh}), this yields
\begin{equation} \label{GtAh2}
    \widetilde A =  G + \sum_\alpha  h_\alpha m_\alpha - \sum_i x_i (E_i - E_i^\ominus),
\end{equation}
since $h_\alpha^\ominus = 0$ by construction, see Eq.~(\ref{Abasic}). In view of Eq.~(\ref{FGm}), the above equation implies:
\begin{equation} \label{AtAh}
    \widetilde A =  A - \sum_i x_i (E_i - E_i^\ominus),
\end{equation}
which generalizes Eq.~(\ref{FtA}), at $M=2^N$, to a broader range of polarization patterns. By combining Eqs.~(\ref{GtAh2}) and (\ref{GtG}), one obtains 
\begin{equation} 
    \widetilde A = \widetilde G + \sum_\alpha h_\alpha m_\alpha \label{AGt}, 
\end{equation}
c.f. Eqs.~(\ref{FGm}). Finally, we take differentials of Eqs.~(\ref{GtG}) and (\ref{AtAh}), respectively, and use Eqs.~(\ref{dGdh}) and (\ref{dGdE}) to see that
\begin{align} 
%    \begin{split}
         m_\alpha &= -\frac{\partial \widetilde G}{\partial h_\alpha} \label{mGth} \\
        h_\alpha &= \frac{\partial \widetilde A}{\partial m_\alpha} \label{hAtm}. 
%    \end{split}
\end{align} 

Eqs.~(\ref{GtAh2}) through (\ref{hAtm}) thus complete the formal task of building a conjoint thermodynamic potential for the coupling constants, on the one hand, and for the expectation values of the degrees of freedom, on the other hand. The thermodynamic potentials $\widetilde A$ and $\widetilde G$ can be thought of as constrained free energies of the system, where the constraint is due to deviations from the standard model. We see that the constrained versions of the Helmholtz and Gibbs energies are in the same relation to each other as their unconstrained counterparts.% c.f. Eqs.~(\ref{dGdh}) and (\ref{hFM}).

Relation (\ref{hFM}) can be profitably thought of as specifying stationary points of a tilted free energy surface $A(\vec m) - \sum_\alpha h_\alpha m_\alpha$ as a function of the magnetizations $m_\alpha$, where the fields $h_\alpha$ are treated as constants representing externally imposed fields. If there is only one stationary point, it corresponds to a unique minimum; the depth of the minimum is then equal to the equilibrium Gibbs energy $G = A - \sum_\alpha h_\alpha m_\alpha$. If, however, a portion of the surface $A(\vec m)$ happens to exhibit a negative curvature, then there is a way to tilt the latter surface so that it will now exhibit more than one minimum. If such additional minima are in fact present---implying Eq.~(\ref{hFM}) has multiple solutions---one is no longer able to unambiguously define the Gibbs energy or other thermodynamic quantities including the entropy and energy. Consequently, the coarse-graining scheme itself becomes ambiguous. These notions apply equally to the constrained free energies $\widetilde A$ and $\widetilde G$, per Eqs.~(\ref{AGt})-(\ref{hAtm}),  which is central to the discussion of reduced descriptions in Section~\ref{reduced}. 

Eqs.~(\ref{weightsh}) and (\ref{G}) also embody a particular way to implement specific inquiries in the present formalism. Suppose, for the sake of argument, that we wish to recover a pattern from a cue in the form of a subset of spins each fixed in a certain direction, similarly to how one would retrieve in the Hopfield network~\cite{Hopfield1982}. This would formally correspond to setting $h_\alpha = \pm \infty$, the two options corresponding to spin $\alpha$ polarized up and down, respectively. More generally, the setup in Eq.~(\ref{G}) allows one to impose a broad variety of constraints, whose rigidity can be tuned by varying the magnitude of the pertinent source field.

\section{\label{reduced} Reduced Descriptions Break Ergodicity} 

The number of coupling constants one can employ in practice is much less than the number $2^N$ of the parameters  constituting the full description from Eq.~(\ref{JlnZ}). This implies that practical descriptions are reduced essentially by construction. Here we argue that the task of finding such reduced, practical descriptions is, however, intrinsically ambiguous. The Section is organized as follows. First, we attempt to emulate a very simple dataset using a trial generative model that is missing a key interaction from the actual model underlying the dataset. We will observe that, in contrast with the complete description, the reduced free energy surface $\widetilde A(\bm J)$ is no longer single-minimum but, instead, develops competing minima. We next show that the Helmholtz free energy of the reduced model {\em also} develops competing minima; these minima correspond exactly with the minima of the potential $\widetilde A(\bm J)$. This connection is, then, used to show that reduced descriptions transiently break ergodicity, which will act to stymie both training and retrieval. A potential remedy will be proposed. 

\begin{figure}[t]
\centering
\includegraphics[width=\figurewidth]{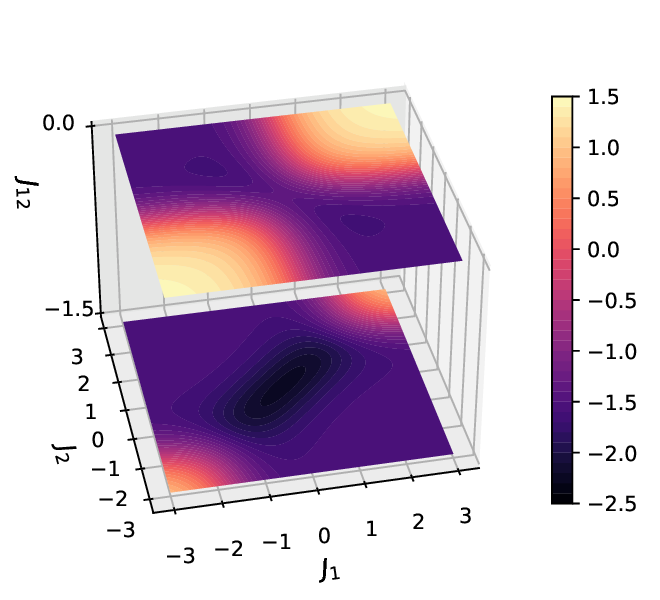}
\caption{\label{AE2fig} The free energy surface $\widetilde A(J_1, J_2, J_{12})$ from Eq.~(\ref{Atilde}), as a function of the coupling constants, for the standard model (\ref{ccE2}); $M=2^N=4$. Two select cross-sections $J_{12} = \text{const}$ are shown. The $J_{12} = -J$ cross-section contains the global minimum of $\widetilde A$, while the $J_{12} = 0$ cross-section corresponds to a reduced description in which the two-body interaction is missing.  $T = T^\circ=1$, $J=1.5$.}
\end{figure}

\begin{figure}[t]
\centering
\includegraphics[width=\figurewidth]{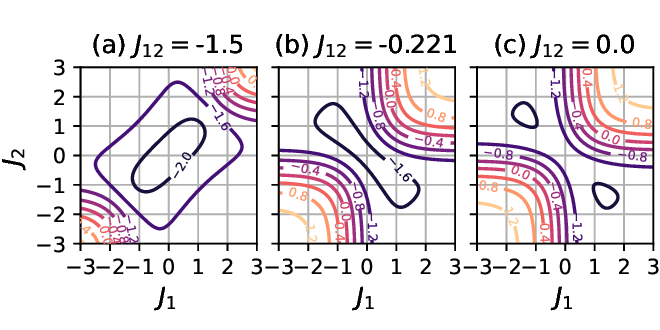}
\caption{\label{AE2figContour} Panels (a) and (c) provide contour plots corresponding to the $J_{12} = 0$ and $J_{12} = - J = -1.5$ slices, respectively, from Fig.~\ref{AE2fig}. Panel (b) exemplifies a special intermediate situation, where the surface $\widetilde A(J_1, J_2, J_{12})$ just begins to develop two distinct minima. $T = T^\circ=1$, $J=1.5$.}
\end{figure}

The simplest system where reduction in description can be performed is a set of two interacting spins. In the full description, there are three independent weights $x_i$; the corresponding generative model will generally employ all three coupling constants, namely, the one-particle fields $J_1$ and $J_2$, and the two-particle, spin-spin coupling $J_{12}$. To this end, let us consider the function $\widetilde A$ for the following standard model:
\begin{equation} \label{E2}
    E^\ominus(\vec \sigma_i) = J \sigma_1 \sigma_2
\end{equation}
where note we use the plus sign in the r.h.s., in contrast with Eq.~(\ref{EGM}). We will assign the distribution $Z_i$ according to the Boltzmann prescription for all descriptions:
\begin{equation}
    Z_i = e^{-E(\vec \sigma_i)/T^\circ}.
\end{equation}
The positive constant $T^\circ$ can be thought of as some normal temperature. A positive $J$ corresponds to a fuzzy-logic inverter gate, a negative $J$ to a fuzzy-logic identity gate. For concreteness, we will consider $J > 0$, without loss of generality. By construction we set up the standard values as follows: 
\begin{equation} \label{ccE2}
\begin{split}
 J^\ominus_{12} &= - J < 0 \\
J^\ominus_0 &= J^\ominus_1 = J^\ominus_2 = 0. 
\end{split}
\end{equation}
The resulting thermodynamic potential $\widetilde A$, as a function of the couplings $\bm J$, is illustrated in Figs.~\ref{AE2fig} and \ref{AE2figContour} in the form of select cross-sections $J_{12} = \text{const}$. Note the $J_{12} = - J$ cross-section contains the global minimum of the potential $\widetilde A$. 

Because of the combinatorial multiplicity of high-rank couplings, we are particularly interested in such trial generative models that employ the smallest possible number of high-rank couplings. Thus for the two spins, we focus on the cross-section $J_{12}=0$ of the overall free energy surface. As can be seen in Fig.~\ref{AE2fig}, the function $\widetilde A(J_1, J_2; J_{12}=0)$ is minimized not at a unique set of the two remaining variables $J_1$ and $J_2$, but, instead, is bi-stable. The two minima are exactly degenerate. This means the optimal choice of the coupling constants---$J_1$ and $J_2$ in this case---is no longer obvious. Indeed, already an infinitesimal change in the coupling constants of the standard model (\ref{ccE2}) can result in a discrete change in the position of the lowest minimum. 

Suppose, for the sake of argument, that we simplemindedly  accept the $J_1 >0$, $J_2 < 0$ minimum as the result of training and ignore the other minimum. This {\em will} stabilize one of the correct configurations of the gate, $\uparrow \downarrow$, but it will also make the remaining correct configuration $\downarrow \uparrow$ of the gate the least likely one out of the four available configurations, during retrieval. As yet another candidate strategy, suppose we accept not the location of an individual minimum of $\widetilde A(J_1, J_2, J_{12} = 0)$ as the result of training, but, instead, use a Boltzmann-weighted average over the minima. For the generative model (\ref{ccE2}), this would result in accepting $J_1 = J_2 = 0$ as the result of training, a nonsensical outcome.

Similar ambiguities arise in statistical descriptions of phase co-existence, when the free energy surface of the system exhibits more than one minimum. We can see this explicitly by expressing the free energy $\widetilde A$ of the reduced description as a function of coarse-grained variables. For the reduced description, we will use the following trial energy function:
\begin{equation} \label{MFansatz}
    E = - \sum_\alpha J_\alpha \sigma_\alpha.
\end{equation}
which corresponds to a widely used meanfield ansatz introduced early on by van der Waals~\cite{LLstat}. Hereby one replaces instantaneous values of the force acting on a given degree of freedom by an effective force that depends only on the configuration of the latter degree of freedom. Thus expectation values of products of quantities pertaining to distinct degrees of freedom factorize as follows:
\begin{equation} \label{MFconstraint}
   \overline{ \prod_\alpha \sigma_\alpha} = \prod_\alpha \overline \sigma_\alpha = \prod_\alpha m_\alpha
\end{equation}
According to Eq.~(\ref{MFconstraint}), the aforementioned effective force acting on a given spin is determined by using the typical values of the magnetization of the other spins, not the actual values.

If one sets all the coupling constants of order two and higher to zero in the generative model (\ref{EGM})---in order to reduce the description or otherwise---the latter model becomes equivalent to the meanfield ansatz (\ref{MFansatz}). This circumstance allows us to use the same notation, viz., $\widetilde A$ for the thermodynamic potential as a function of either the coupling constants $J_i$, on the one hand, or the magnetizations $m_i$, on the other hand.

Let us rewrite Eq.~(\ref{FtA}), at $M=2^N$, to yield the following expression:
\begin{equation} \label{Atilde4}
    \widetilde A = \sum_i x_i E_i^\ominus  - T S,
\end{equation}
where $S$ is the conventional canonical entropy of the trial model. Indeed, $A = \sum_i x_i E_i - T S$. According to Eq.~(\ref{Atilde4}), $\widetilde A$ for a trial generative model can be computed by first writing down the expression for the free energy of the trial energy function and, then, simply replacing the energy part of the expression by the expectation value of the actual expression for the energy, however evaluated using the Boltzmann weights of the trial ansatz. 

The entropy for a standalone spin $\sigma=\pm 1$ is simply the mixing entropy of a binary mixture with mole fractions $(1+m)/2$ and $(1-m)/2$, respectively. In view of Eq.~(\ref{Atilde4}), then, the Helmholtz free energy for the most general energy function (\ref{EGM}) computed under the meanfield constraint (\ref{MFconstraint}) reads:
\begin{eqnarray} \label{FMF}
    \widetilde A  =  E^\ominus (\vec \sigma)|_{\vec \sigma = \vec m}   +  T \sum_\alpha \left[\frac{1+m_\alpha}{2} \ln \frac{1+m_\alpha}{2} + \frac{1-m_\alpha}{2} \ln \frac{1-m_\alpha}{2} \right]. 
\end{eqnarray}
where the expression $E^\ominus (\vec \sigma)|_{\vec \sigma = \vec m}$ refers to the energy function (\ref{EGM}) evaluated using the standard values of the coupling constants and setting $\sigma_\alpha \to m_\alpha$. For example, for two spins, the above expression becomes
\begin{eqnarray} \label{FMF2}
    \widetilde A  & = &  -J_0^\ominus - J_1^\ominus m _1 - J_2^\ominus m_2 - J_{12}^\ominus m_1 m_2  \nonumber \\ & + & T \sum_{\alpha=1}^2 \left[\frac{1+m_\alpha}{2} \ln \frac{1+m_\alpha}{2} + \frac{1-m_\alpha}{2} \ln \frac{1-m_\alpha}{2} \right].
\end{eqnarray}

\begin{figure}[t]
\centering
\includegraphics[width=\figurewidth]{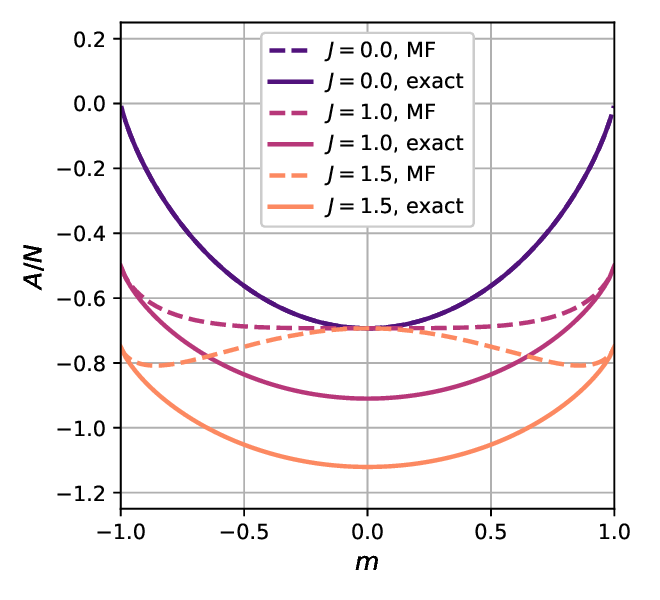}
\caption{\label{Am1m2Figslice} Dashed lines: the $m \equiv m_1 = - m_2$ slice of the free energy surface $\widetilde A(m_1, m_2)$ from Eq.~(\ref{FMF3}) for three select values of $J$.  Solid lines: the respective values of the exact Helmholtz energy $A$ for the energy function (\ref{E2}). $T=T^\circ=1$.}
\end{figure}

Let us examine consequences of using the meanfield expression (\ref{FMF2}), as a trial description, specifically for the two-spin system from Eq.~(\ref{E2}). Thus we set $J_0^\ominus=J_1^\ominus=J_2^\ominus=0$ and $J_{12}^\ominus=-J$:
\begin{eqnarray} \label{FMF3}
    \widetilde A   =  J m_1 m_2   +  T \sum_{\alpha=1}^2 \left[\frac{1+m_\alpha}{2} \ln \frac{1+m_\alpha}{2} + \frac{1-m_\alpha}{2} \ln \frac{1-m_\alpha}{2} \right].
\end{eqnarray}
The interaction term $J m_1 m_2$ is that of an antiferromagnet. The free energy surface is readily seen to become bi-stable for $J/T>1$, implying the meanfield magnet (\ref{FMF3}) exhibits a criticality in the form of a paramagnet-to-antiferromagnet transition with a N\'eel temperature $T_C = J$. In contrast, the {\em exact} free energy is single-minimum for all values of $J$, implying the meanfield solution becomes qualitatively incorrect for $J/T>1$. In Fig.~\ref{Am1m2Figslice}, we show the slice of the free energy surface along the $m \equiv m_1 = - m_2$ line, for three select values of $J$. For reference, we also show the corresponding values of the exact free energy, which can be readily evaluated numerically using Eqs.~(\ref{G}), (\ref{dGdh}), and (\ref{FGm}). 

The two minima on the meanfield  curve for $J=1.5$, Fig.~\ref{Am1m2Figslice}, are precisely equivalent to the two minima of $\widetilde A(J_1, J_2, J_{12})$ in the $J_{12}=0$ plane in Fig.~\ref{AE2fig}, because the generative model (\ref{EGM}) at $J_{12}=0$ (at $N=2$) is equivalent to the meanfield ansatz (\ref{MFansatz}), as already alluded to.

According to the discussion following Eq.~(\ref{hFM}), we do not expect the thermodynamics of the setup (\ref{FMF3}) to be well defined below the critical point, $J/T > 1$, where the surface (\ref{FMF3}) has more than one minimum.  In physical terms, the nominal expectation value of the magnetization---which remains vanishing even below the critical point---now becomes decoupled from the set of its typical values and, thus, is no longer descriptive of the microscopic behavior of the system. In formal terms, the corresponding Gibbs energy, the entropy, and the energy all become poorly defined and can be thought of, at best, as multi-valued under limited circumstances. We flesh out these notions in what follows; the detailed calculations are relegated to Appendix~\ref{spin2calculation}. 

\begin{figure}[t]
\centering
\includegraphics[width=\figurewidth]{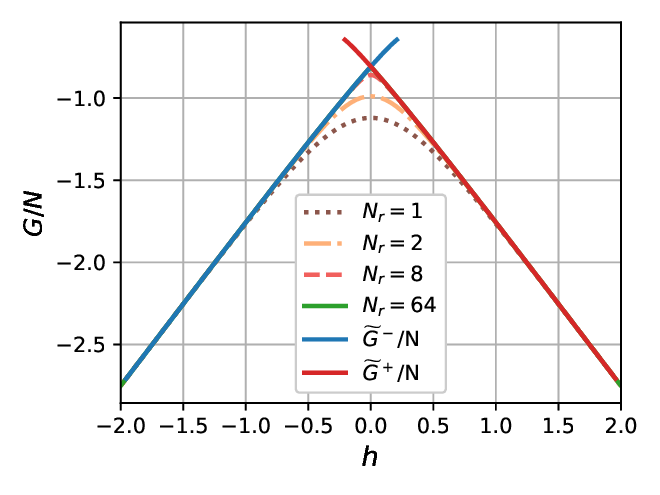}
\caption{\label{twovaluedFig} The two branches of the restricted Gibbs energy $\widetilde G$ are labeled with $\widetilde G^-$ and $\widetilde G^+$. The equilibrium Gibbs energy $\widetilde G_\text{eq}$ from Eq.~(\ref{Gcg}) is shown for select values of $N_r$. The $N_r=64$ curve is masked by the meanfield curves. $J=1.5$, $T=T^\circ=1$.}
\end{figure}

\begin{figure}[t]
\centering
\includegraphics[width=\figurewidth]{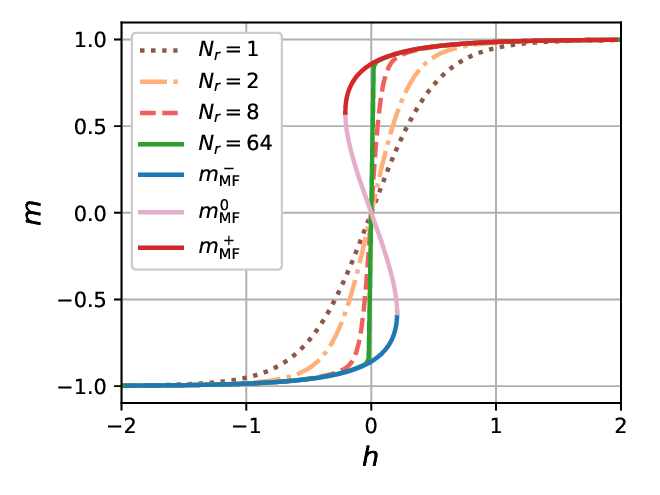}
\caption{\label{mcan} The magnetization, per spin, as a function of the source field along the slice $m=m_1=-m_2$, $h=h_1=-h_2$, below the critical point.  The meanfield curves correspond to the $m_1 > 0$, $m_2 < 0$ minimum (red), the $m_1 < 0$, $m_2 > 0$ minimum (blue), and the mechanically unstable region separating the spinodals (pink); the portions of this curve where $mh <0$ are either metastable or unstable. The equilibrium values for the energy function (\ref{Exz}) are given for several system sizes. $T=T^\circ=1$, $J=1.5$.}
\end{figure}

Having two distinct minima on the $\widetilde A(m)$ curve implies the Gibbs energy is no longer well defined. Indeed, on the one hand the typical values $m_\alpha$ are non-vanishing below the critical point already at $h_\alpha =0$. On the other hand, the average magnetization should be zero, at $h_\alpha =0$, by symmetry. This seeming paradox is resolved by noting that symmetry-breaking fields acting on {\em individual} spins do not have to be externally imposed, but could emerge internally \cite{Goldenfeld, L_AP}; such fields originate from the surrounding spins and come about self-consistently. One may then define a multi-valued Gibbs energy, the number of branches determined by the degree of degeneracy of the Helmholtz energy. The two branches of the Gibbs energy---one per each minimum of $\widetilde A(m)$---are shown in Fig.~\ref{twovaluedFig} and labeled ``$\widetilde G^+$'' and ``$\widetilde G^-$'' as pertinent to the r.h.s. and l.h.s. minimum of $\widetilde A$, respectively. Each of the two branches thus corresponds to a restricted Gibbs energy computed for a single phase. Indeed, the differential relation in Eq.~(\ref{hAtm}) can sample the Helmholtz energy only within an individual minimum. Sampling the phase space within an individual minimum can be thought of as vibrational relaxation, a relatively fast process~\cite{HLtime}. In contrast, crossing  over to the other minimum requires a discontinuous transition across the mechanically unstable region delineated by the spinodals, see Figs.~\ref{mcan} and \ref{mmhhFig}. In terms of kinetics, such discontinuous transitions are slow because they require activation~\cite{HLtime}, see also below.

On the other hand the {\em equilibrium} Gibbs energy from Eq.~(\ref{G}) corresponds, by construction, to statistics that are gathered on times longer than any relaxation times in the system, including, in particular, the times of activated escape from individual free energy minima. To compute the equilibrium Gibbs energy corresponding to the meanfield ansatz, we need to implement the constraint (\ref{MFconstraint}) in the canonical sum over the configurations in Eq.~(\ref{G}), where the energy function is specifically from Eq.~(\ref{E2}). To do so, we first recall that the canonical construction implies we are considering a very large number $N_r$ of equivalent non-interacting replicas of the system so as to determine the likeliest mixture of the states of the individual replicas~\cite{McQuarrie}.  Denote the replicas of spin 1 as $\xi_q$, $q = 1, \ldots, N_r$ and the corresponding replicas of spin 2 as $\zeta_q$, $q = 1, \ldots, N_r$. Each pair $(\xi_q, \zeta_q)$ can be thought of as the configuration of the spin pair $(\sigma_1, \sigma_2)$ in sample $q$. The energy per replica is, then, $E/N_r = \left( \sum_q^{N_r} \xi_q \zeta_q \right)/N_r$. To implement the constraint (\ref{MFconstraint}), we need to decouple the fluctuations of spin $2$ from the fluctuations of spin $1$. To accomplish this, we replace, in the correct sum $\sum_q^{N_r} \xi_q \zeta_q$, the actual value of $\xi_q$ by its average value: $\sum_q^{N_r} \xi_q \, \zeta_q \to  \sum_q^{N_r} (\sum_s^{N_r} \xi_s/N_r) \, \zeta_q$. As a byproduct of the latter replacement, spin $2$ now {\em also} enters as an average over the replicas and so no further averaging is needed. Thus we arrive at the following energy function:
\begin{equation} \label{Exz}
    E(\xi, \zeta) = \frac{J}{N_r} \left( \sum_s^{N_r} \xi_s \right) \left( \sum_q^{N_r} \zeta_q \right). 
\end{equation}
We see the meanfield constraint effectively means the replicas in the canonical ensemble now interact with each other. The corresponding equilibrium Gibbs energy reads as follows:
\begin{equation} \label{Gcg}
    \widetilde G_\text{eq} = - T \ln \sum_{\xi_s, \zeta_q} e^{- \frac{E(\xi, \zeta)}{T}   + \frac{h_1}{T} \left( \sum_s^{N_r} \xi_s \right) + \frac{h_2}{T} \left( \sum_q^{N_r} \zeta_q \right)}. %\nonumber \\ &= - T \ln \left( \int \frac{d \lambda}{(2 \pi T/NJ)^{1/2}} e^{N \left\{ \ln 2 \cosh \left[ \frac{\lambda J + h}{T} \right]  - \frac{J^2 \lambda^2}{2} \right\} } \right).
\end{equation}

\begin{figure}[t]
\centering
\includegraphics[width=\columnwidth]{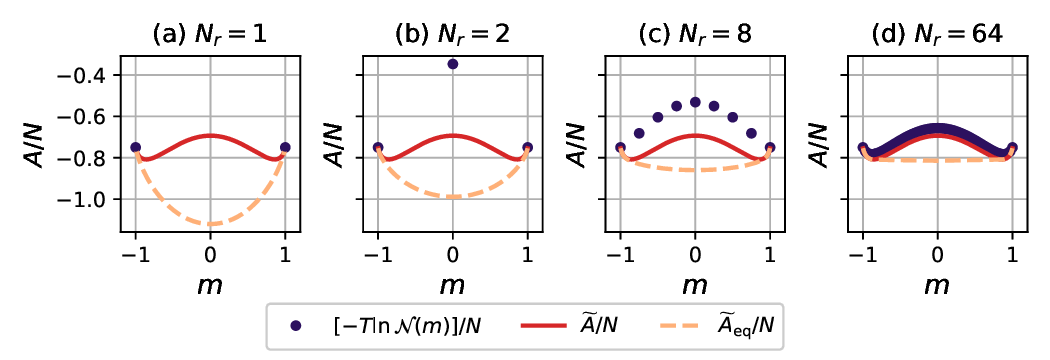}
\caption{\label{Fcan} The equilibrium Helmholtz free energy $\widetilde A_\text{eq}$ for the energy function (\ref{Exz}) compared with the actual free energy cost, $- T \ln {\cal N}(m)$, to have  magnetization $m$ and the meanfield Helmholtz free energy $\widetilde A$. The horizontal axis refers to the magnetization along the $(1, -1)$ direction as in Fig.~\ref{Am1m2Figslice}. $T=T^\circ=1$, $J=1.5$.}
\end{figure}

At its face value, the energy function (\ref{Exz}) corresponds to a set of $N=2 N_r$ spins that interact via a very long-range interaction, whereby an individual spin interacts with a half of the spins. Still, the interaction depends on the system size $N$ in such a way that the total energy  scales linearly with $N$; thus it is a proper, extensive energy. Notwithstanding the somewhat artificial nature of the interaction, the energy function (\ref{Exz}) should actually be a decent approximation for the energetics of a small magnetic domain. Such small domains happen to break ergodicity on sufficiently long times to be useful for information storage; thus their free energy must exhibit at least two free energy minima already for modest values of $N$.  We show that just this is the case for the energy function (\ref{Exz}) in what follows; we will compute the equilibrium values of the basic thermodynamic quantities alongside. Everywhere below, the extensive quantities are shown per spin. We continue focusing on the regime below the critical point, specifically at $J=1.5$. We show equilibrium values of the magnetization $m(h)$ in Fig.~\ref{mcan} and the corresponding equilibrium Gibbs energy $\widetilde G_\text{eq}(h)$ in Fig.~\ref{twovaluedFig}, for select values of $N = 2 N_r$. This, then, allows us to compute, via the Legendre transform, the equilibrium Helmholtz free energy $\widetilde A_\text{eq}(m)$, shown with the dashed line in Fig.~\ref{Fcan}. Also in Fig.~\ref{Fcan}, we  show the actual free energy cost $- T \ln {\cal N}(m)$ to have magnetization $m = m_1 = - m_2$, where 
\begin{equation}
  {\cal N} (m) \equiv \sum_{\{\xi_s, \zeta_q\}} e^{-E(\xi, \zeta)/T} \: \delta_{\sum_s \xi_s, N_r m} \: \delta_{\sum_q \zeta_q, -N_r m}
\end{equation}
Here, $\delta$ is the Kronecker delta and the quantity $m$ admits by construction a discrete set of equally-spaced values, the number of distinct values equal to $(N_r+1)$.

We see the equilibrium free energies each provide a lower bound on the restricted free energies---convex-up for Gibbs and convex-down for Helmholtz---as they should~\cite{LLstat}. Thus, when the distribution of the magnetization becomes bimodal, the {\em equilibrium} Helmholtz energy no longer reflects the actual distribution. Instead, the central portion of the equilibrium Helmholtz energy $\widetilde A_\text{eq}$ becomes a shallow bottom that largely interpolates between the two minima of the restricted Helmholtz energy $\widetilde A(m)$. The meanfield free energy $\widetilde A$, on the other hand, has a qualitatively correct form and, we see, tends asymptotically to the actual free energy cost $- T \ln {\cal N} (m)$ for large $N$, as it should. Transitions between the two free energy minima of $\widetilde A(m)$ are accompanied by an extensive change of total magnetization $\sum_s \xi_s \sim (- \sum_q \zeta_q) \propto N$ and, thus, can be induced by varying the external field by a small amount that scales as $1/N$; this notion is consistent with Fig.~\ref{mcan}. Consequently, the variation of the equilibrium Helmholtz energy, per particle, along its shallow bottom scales as $1/N$ with the system size $N$. Thus, it vanishes in the thermodynamic limit $N$. Consequently, the equilibrium susceptibility becomes poorly defined even though the susceptibilities of individual phases each remain well defined and finite. 

\begin{figure}[t]
\includegraphics[width=\columnwidth]{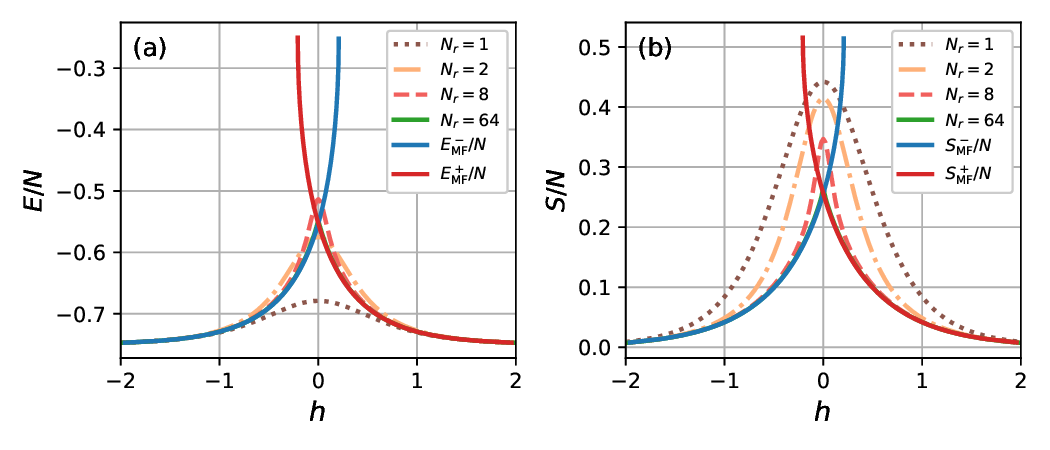}
\centering
\caption{\label{SEhFig} The internal energy (panel (a)) and the entropy (panel (b)). The meanfield energy and entropy, respectively, corresponding to the r.h.s. and l.h.s. minimum of $\widetilde A(m)$ in Fig.~\ref{Fcan} are labeled with supserscripts ``$+$'' and ``$-$.'' $J=1.5$. The equilibrium energy and entropy, respectively, are labeled using the value $N_r$ they were computed for. (The $N_r = 64$ curves are masked by the meanfield curves.)}
\end{figure}

Thus, when ergodicity is broken, the notions of free energy and thermodynamic quantities become poorly defined, c.f. the discussion following Eq.~(\ref{hFM}). We show the restricted and equilibrium values of the internal energy and the entropy in panels (a) and (b), respectively, of Fig.~\ref{SEhFig}. These graphs directly convey that already for a  generative model as simple as the one in Eq.~(\ref{E2}), a reduced description does not allow for a Helmholtz machine~\cite{DayanHintonNealEtAl95}, since the basic thermodynamic quantities, such as the energy and entropy, cannot be properly defined. 

In a finite system, the barriers separating distinct free energy minima are finite, even if tall, implying the ergodicity is broken {\em transiently}. The ergodicity is restored on times that scale exponentially with the parameters of the problem and can become very long already for modestly sized systems and/or following small temperature variations. We illustrate these notions in Appendix~\ref{transient}. There we also demonstrate that transitions among distinct free energy minima are rare events that occur via non-generic sequences of individual spin flips. 

Below the critical point, the minima of the surface $\widetilde A$ are two and are separated by a finite barrier. Thus, the machine will be able to reproduce only the patterns belonging to the now reduced phase space. Specifically which minimum will be found during training is generally a matter of chance; the odds are system-specific. For instance, setting $J_1^\ominus$ at a small positive value, in the generative model (\ref{ccE2}), will create a bias toward the $J_1 >0$, $J_2 < 0$ minimum of the $\widetilde A(J_1, J_2)$ surface. One can use symmetry considerations, see Appendix~\ref{shape}, to make a general case that the reduced free energy surface will consist of isolated minima whose curvature is non-vanishing in all directions. This means that the bound states corresponding to the latter minima are compact, shape-wise, in the phase space.

The two minima on the surface $\widetilde A(m_1, m_2)$, for the model (\ref{E2}), correspond to the two ``correct'' states of the inverter gate, $\uparrow \downarrow$ and $\downarrow \uparrow$, respectively. Both minima must be sampled in order for the machine to work adequately. Owing to the aforementioned one-to-one correspondence between the minima of $\widetilde A(J_1, J_2)$ and $\widetilde A(m_1, m_2)$, respectively, the two minima on the $\widetilde A(J_1, J_2)$ surface must be likewise sampled for the machine to work properly. Sampling multi-modal distributions can become computationally unwieldy already for a modest number of variables. Thus to efficiently sample multiple free energy  minima that are separated by barriers, it may be necessary to employ a separate generative model---call it an ``expert''---for each individual minimum. For the underlying model (\ref{E2}), one such expert---in the form of fields $J_1 > 0$ and $J_2 < 0$ in Eq.~(\ref{MFansatz})---can be used to retrieve the $\uparrow \downarrow$ state of the gate, as already mentioned.  The other model---in the form of fields $J_1 <0$ and $J_2 >0$---will retrieve the $\downarrow \uparrow$ state. 

The requisite number of experts is thus bounded from above by how many minima free energy landscape will acquire as a result of reduction in description. Actual physical systems provide insight as to how large this number can be. Note that already lowering the temperature in a physical system raises the free energy and, thus, corresponds to a reduction in description. There are a variety of physical scenarios that can unfold as a set of interacting degrees of freedom is cooled. For instance, translationally invariant Ising-spin models with ferromagnetic, short-range coupling will exhibit two minima below the ergodicity breaking transition, irrespective of the system size. The worst-case scenario is arguably represented by some disordered spin systems, such as the meanfield Potts glasses~\cite{MCT1}, in which the number of minima can scale exponentially with the system size in the broken-ergodicity regime. The disorder can also be self-generated---as in actual finite-dimensional glassy liquids---whereby the free energy surface becomes exponentially degenerate below the dynamical crossover~\cite{L_AP, LW_ARPC}. In the latter case, a reduced description has in fact been achieved \cite{dens_F1, BausColot, RL_LJ, L_AP} using a trial density profile of a glassy liquid in the form of a sum of narrow Gaussian peaks each centered at a distinct corner of an aperiodic lattice. The number of distinct, comparably stable aperiodic structures self-consistently turns out to scale exponentially with the system size, consistent with experiment \cite{XW, L_AP, LW_ARPC}. Each such lattice is mechanically stable and corresponds to a distinct expert.

We provide an elementary illustration in Appendix~\ref{shape} that the greater the reduction in the description, the more experts are needed for proper functionality. Conversely one may say that an individual expert operating on a progressively naive model will retrieve a smaller subset of the correct configurations. 
%The liquid density ansatz from \cite{dens_F1, BausColot, RL_LJ, L_AP} is both one-particle and meanfield---corresponding to the Einstein-oscillator approximation; it is entirely analogous to the meanfield ansatz (\ref{MFansatz}). 
In any event, we have described elsewhere~\cite{HeThesis} a protocol to train sets of experts specifically for binary datasets; these findings are also to be presented in a forthcoming submission. 

\section{\label{falsehood} Knowledge as a Library of Bound States}

The number of spins $N$ in a description depends on the resolution. Even at relatively lower resolutions and, hence, modest values of $N$, the number of configurations represented in a realistic dataset will be much much less than the size $2^N$ of the phase space available to $N$ binary variables. Consequently, the vast majority of the energies in the full description (\ref{JlnZ}) are undetermined.  Here we argue that the energies of non-represented configurations must be explicitly parameterized and that such parameterizations, as well as retrieval protocols, must satisfy rigid constraints in order to avoid instabilities toward learning or retrieval of non-represented configurations. One may think of non-represented configurations as false positives. 

Denote the number of configurations that are in fact represented in the dataset with $M_0 < 2^N$ and number them using $i = 1, \ldots, M_0$.  For the sake of argument, we will sometimes call the represented configurations ``true,'' while referring to the non-represented configurations as ``false:''
\begin{eqnarray}
    \text{represented (``true'')} &: & 1 \le i \le M_0 \label{true} \\
    \text{non-represented (``false'')} & : &  M_0 < i \le 2^N \label{false}
\end{eqnarray}

We stipulate, for concreteness, that the model performs satisfactorily, if the total weight of the false states, relative to the total weight of the true states, be less than one:
\begin{equation} \label{pfpt}
    \frac{P_\text{F}}{P_\text{T}} = \frac{Z_\text{F}}{Z_\text{T}} \le 1. 
\end{equation}
Here,
\begin{equation} \label{Zrestricted0}
\begin{split}
    Z_\text{T} & \equiv \sum_{i=1}^{M_0} Z_i \\
    Z_\text{F} & \equiv \sum_{i=M_0+1}^{2^N} Z_i. 
\end{split}
\end{equation}
Inequality (\ref{pfpt}) should be thought of as imposing an upper bound on the probability to produce a false positive. The actual numerical value of the bound on the r.h.s. of Eq.~(\ref{pfpt}) is not essential for what follows.

The contributions of the non-represented reconfigurations to the sums in Eq.~(\ref{JlnZ}) are indeterminate, as already alluded to. Simply omitting the latter contributions from the sums in Eq.~(\ref{JlnZ}) would be formally equivalent to assigning a {\em fixed}, vanishing value to their energies $E_i=0$, irrespective of the detailed parameterization of the represented states. Here, instead, we take a general approach whereby we treat the energies of the non-represented states as adjustable parameters; their values, then, will be chosen so as satisfy the constraint in Eq.~(\ref{pfpt}). 

For concreteness, we choose a calibration where $Z_i^\ominus$ is set to the number of times configuration $i$ was presented in the dataset. (For those many configurations that were {\em not} presented in the dataset, we treat the corresponding $Z_i^\ominus$ as adjustable parameters.)  Also for concreteness, we connect the standard weights and energies, respectively, using a Boltzmann weight-like relation, as in Section~\ref{reduced}:
\begin{equation} \label{ElnZ}
    E_i^\ominus = - T^\circ \ln Z_i^\ominus.
\end{equation}
where $T^\circ$ is a positive constant. It is understood that some of the ``true'' states can be false positives, while some of the ``false'' states can be false negatives. For this reason, the two respective sets of energies will generally overlap. Also for the same reason, we will employ a functional relation $E_i=E_i(\{ Z_j \})$ that is not specific to configurations being represented or not, so as not to prejudge the veracity of a pattern on the basis of its being present or absent in any given dataset. In any event, the parameterized energy set (\ref{ElnZ}) unambiguously defines a generative model of the type in  Eq.~(\ref{EGM}), according to Eq.~(\ref{JlnZ}).

Owing to the huge number of configurations, the criterion (\ref{pfpt}) can be profitably recast in free-energetic terms, which is our next step. We have for the restricted partition functions of the true and false states, respectively:\begin{align} 
%\begin{split}
    Z_\text{T}(T) & \equiv \sum_{i=1}^{M_0} e^{-E^\ominus_i/T} \label{ZT} \\
    Z_\text{F}(T) & \equiv \sum_{i=M_0+1}^{2^N} e^{-E^\ominus_i/T}. \label{ZF}
%\end{split}
\end{align}
The individual Boltzmann weights correspond to the weights of respective configuration during Gibbs sampling of the generative model at temperature $T$; thus the quantity $T$ may be designated as the ``retrieval temperature.'' To avoid confusion, we note that the ensemble in Eqs.~(\ref{ZT}) and (\ref{ZF}) is the conventional canonical ensemble, not the ensemble (\ref{ZE}). The energies $E_i$ are kept strictly constant, while the temperature $T$ does not set energy units but, instead, is allowed to take any positive value by construction.  

To the partition functions (\ref{ZT}) and (\ref{ZF}), there correspond restricted Helmholtz free energies:\begin{align} 
    A_\text{T}(T) = - T \ln Z_\text{T}(T) \\
    A_\text{F}(T) = - T \ln Z_\text{F}(T),
\end{align}
The condition (\ref{pfpt}) is then equivalent to stipulating that the true states are thermodynamically stable relative to the false states:
\begin{equation} \label{AA}
    A_\text{T} (T) \le A_\text{F} (T).
\end{equation}
To satisfy this constraint, two conditions must be met: (1) There should be a temperature interval, in which the Helmholtz energy of the true states is lower than the Helmholtz energy of the false states. (2) Retrieval must be performed at a temperature from the latter interval. To establish the implications of these constraints for the  parameterization of the non-represented states, we first introduce the canonical energy and entropy:\begin{align}
 E(T) &= - \frac{\partial \ln Z}{\partial (1/T)} \label{Ecan} \\
 S(T) &= \frac{E-A}{T} \label{Scan}
\end{align}
where the partition function $Z$ can be either one of the restricted partition functions from Eqs.~(\ref{ZT}) and (\ref{ZF}) or the full, equilibrium partition function
\begin{equation} \label{Zeq}
    Z_\text{TF} \equiv Z_\text{T} + Z_\text{F}.
\end{equation}

In a usual way~\cite{LLstat}, Eqs.~(\ref{Ecan}) and (\ref{Scan}) uniquely specify a coarse-grained log-spectrum $S(E)$, with the variable $T$ being the parameter, for any system with a positive heat capacity. %Indeed, under these circumstances $S(E)$ is a strictly convex-up function: $\partial^2 S/\partial E^2 = -1/T^2 C < 0$. 
Conversely, given a temperature $T$, the canonical energy $E$ of a single phase system is determined by the location of the tangent to the curve $S(E)$ whose slope is equal to $1/T$:
\begin{equation} \label{Tmicro}
     \frac{\partial S}{\partial E} = \frac{1}{T}. 
\end{equation}

%We next adopt a specific parameterization One can obtain the parametric curves $S(E)$ for the individual phases, using the canonical $E(T)$ and $S(T)$ from Eqs.~(\ref{Ecan}) and (\ref{Scan}); these are shown in Fig.~\ref{blocks}. 

We assume for now that the individual spectra of the true and false states, respectively, each exhibit a positive heat capacity throughout. The energies $\{ E_i^\ominus \}$ of the model from Eq.~(\ref{ElnZ}) are defined in terms of the positive constant $T^\circ$. This automatically fixes the ratio $E(T^\circ)/T^\circ$ for the restricted and equilibrium partition functions alike, given a specific form of the distribution of the dimensionless quantities $\{ E_i^\ominus/T^\circ \}$; in other words, the constant $T^\circ$ simply specifies the energy units. Consequently, Eqs.~(\ref{AA}) and (\ref{Scan}) imply: 
\begin{equation} \label{Egapmin}
    [E_\text{F}(T^\circ) - E_\text{T}(T^\circ)]/T^\circ \ge S_\text{F}(T^\circ) - S_\text{T}(T^\circ).
\end{equation}
One can also infer the above inequality directly from Eqs.~(\ref{pfpt}) and (\ref{ElnZ}) by noting that $E/T^\circ = - \sum_i Z_i^\ominus \ln Z_i^\ominus/\sum_j Z_j^\ominus$ and $S = - \sum_i Z_i^\ominus \ln( Z_i^\ominus/\sum_j Z_j^\ominus)/\sum_k Z_k^\ominus$
where the summations are consistently over either the true or the false set of states,  respectively. 

Eq.~(\ref{Egapmin}) dictates that the canonical energy of the false states in a robust description must be separated by a non-vanishing, extensive {\em gap} from the canonical energy of the true states. Indeed, for $M_0 \ll 2^N$, the r.h.s. of Eq.~(\ref{Egapmin}) is numerically close to $N \ln 2$.  We note that for dataset parameterizations obeying condition (\ref{Egapmin}), the standard entropy (\ref{shannon}) is dominated by the represented states, implying the present treatment is internally-consistent. 

\begin{figure}[t]
    \centering
    \includegraphics[width=\figurewidth]{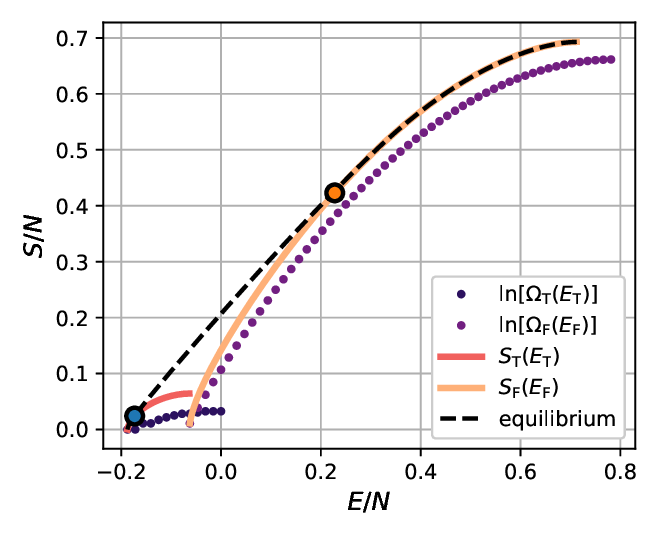}
    \caption{\label{blocks} Parametric graph of the entropy as a function of energy from Fig.~\ref{SET}. The dots show the model, discrete energy distribution $\Omega_\text{T}(E_\text{T})$ and $\Omega_\text{F}(E_\text{F})$ that were used to compute the restricted and the equilibrium quantities, respectively. The blue and orange dot show the locations corresponding to temperature $T= T^\circ$. The location, energy-wise, of the false states was parameterized so that they are in equilibrium with the true states at $T=T^\circ$.  $N=64$. }
\end{figure}

The requisite presence of a non-vanishing, extensive energy gap means that the set of the true states and the set of the false states, respectively, can be thought of as comprising two separate {\em phases}; the energy gap corresponds to the latent heat of the transition. The condition (\ref{AA}) for the generative model to be able to retrieve the represented configuration is thus equivalent to requiring that the phase comprised of the true states be stable relative to phase composed of the false states. Specifically, the true states will be the stable phase at temperatures $T$ such that
\begin{equation} \label{TT0}
    T < T_0,
\end{equation}
where $T_0$ is the (phase-transition) temperature at which the two phases are in mutual equilibrium:  
\begin{equation} \label{T0}
    \frac{1}{T_0} = \frac{S_\text{F}(T_0) - S_\text{T}(T_0)}{E_\text{F}(T_0) - E_\text{T}(T_0)}.
\end{equation} 
Note that $T_0 \ge T^\circ$, in view of Eq.~(\ref{Egapmin}). We illustrate the borderline case $T_0 = T^\circ$ in Fig.~\ref{blocks}, using {\em ad hoc} energy distributions for the true and false states, $\Omega_\text{T}(E_\text{T})$ and $\Omega_\text{F}(E_\text{F})$, respectively. The latter distributions are displayed in the same Figure using the dots, while the corresponding dependences of the canonical entropy on the canonical energy are shown with solids lines. We refer the reader to Appendix~\ref{FTcoexist} for a detailed discussion. There we also explicitly illustrate that each of two phases corresponds to a free energy minimum. 

The thermodynamic singularity at $T_0$ defines a dichotomic distinction between the true and false states, even if their respective spectra overlap. Because of its high entropy, the high-$T$ phase can be thought of as corresponding to a generic mixture of $\sim 2^N$ components. The low-$T$ phase, on the other hand, is by construction a mixture with a dataset-specific composition, whereby some components are represented much more than others. Thus at the temperature $T_0$, the composition of the mixture undergoes an intrinsically {\em discrete} change, consistent with the transition exhibiting a latent heat.

The states within the energy gap $[E_\text{T}(T_0), E_\text{F}(T_0)]$ are strictly inaccessible in equilibrium because at any temperature, there is at least one state whose free energy is lower. This implies that a subset of the true states, viz. those at $E > E_\text{T}(T_0)$, could not be observed in equilibrium. This high-energy flank of the true states corresponds, in the calibration (\ref{ElnZ}), to low-frequency, relatively underrepresented configurations from the dataset. Although the latter states {\em can} be observed, in principle, at temperatures above $T_0$---as we illustrate in Fig.~\ref{AtEFig}(b) in Appendix~\ref{FTcoexist}---the true states, as a whole, are however only {\em meta}stable at such temperatures. If kept at $T>T_0$, the machine will spontaneously transition into the false states---and thus will begin to produce non-represented configurations. The machine is thermodynamically unlikely to retrieve the represented configurations again, unless the temperature is lowered below $T_0$. Thus we view robustness of retrieval as conditional on the machine being confined to the portion of the phase space pertaining to the true states, which implies a breaking of ergodicity. 

Thus the totality of the true states can be thought of as a collection---or library~\cite{LW_aging} if you will---of bound states centered at patterns from the dataset. The bounding potential is a {\em free} energy since it has an entropic component. The spectrum of the non-represented states must be properly parameterized to avoid escape toward non-represented states; such an escape may well occur in an abrupt, avalanche-like fashion owing to the transition being discontinuous. This notion  is consistent with an analysis of US Supreme Court data, due to ~\cite{gresele2017maximum}, who used a spin-based generative model of the type considered here. The latter study associates frequencies of configurations with Boltzmann weights. The non-represented states---called ``unobserved'' in \cite{gresele2017maximum}---are omitted from the sums in the expressions for the coupling constants. According to the discussion in the beginning of this Section, omitting unobserved states amounts to pinning their energies at $E_i=0$; the latter value happens to be greater than the energies assigned to the observed states. \cite{gresele2017maximum} find that including in the dataset configurations whose weight is lower than a certain threshold value causes an instability toward faulty retrieval. The present study suggests that those low-weight states may well be sufficiently close, energy-wise, to the non-represented states so as to substantially stabilize them.

The stability criterion (\ref{TT0}) must be satisfied by descriptions irrespective of whether they employ full or reduced sets of coupling constants. This amounts to an additional constraint, when optimizing with respect to the values of the coupling constants and/or number of experts, in a reduced description. These conclusions are consistent with earlier studies of associative memory Hamiltonians (AMH) for protein folding. AMH-based generative models have been used to predict three-dimensional structures of native folds of proteins since the 1980s~\cite{FriedrichsWolynes1989, doi:10.1021/jp212541y}. The native structures are extracted from protein-structure databases, while the non-native states are emulated by placing residues but generically within correct structures. The coupling constants are then determined by maximizing the energy gap separating the native and non-native states, respectively, relative to the width of the spectrum of the non-native states. %Retrieval corresponds to sampling the Boltzmann distribution corresponding to the energy function. 

A discussion of practical strategies as to finding reasonable reduced sets of coupling constants is beyond the scope of this paper. Still the stability criterion (\ref{AA}) appears to provide some insight. We note that a single spin $\sigma$ or any product of distinct spins $\prod_{i=1}^n \sigma_{\alpha_i}$ ($n \ge 1$) vanishes, if averaged over the full set of $2^N$ states---this is a simple consequence of Eq.~(\ref{JlnZ}). For energy gaps $\Delta E$ significantly greater than the respective widths of the spectra of the true and false states, one may approximately write, according to Eq.~(\ref{JlnZ}):
\begin{equation}
    J_{\alpha_1, \ldots, \alpha_n} \propto  (\Delta E) \la \prod_{i=1}^n \sigma_{\alpha_i} \ra_\text{T}
\end{equation}
up to a multiplicative constant, where the averaging is over the represented states. Thus, one may use Hebbian learning rules for initial screening of important interactions.

Finally we briefly comment on datasets and/or parameterizations thereof where the distribution $\{ Z_i \}$ is multi-modal. In such cases, the true states themselves may be best thought as a collection of distinct phases whose stability relative to each other and to the false states will depend on temperature. When distributed, the numbers $Z_i$ can be profitably thought of as numerical labels; we refer the reader to \cite{marsili2013, haimovici2015, cubero2019} for an in-depth discussion of such labeling schemes.   

\section{\label{sec:conclusion} Summary and Concluding Remarks}

We have considered acquisition of knowledge in the form of a generative model that operates on binary variables. The generative model assigns an energy value to each of the $2^N$ configurations of $N$ such binary variables. The expression for the energy has the functional form of a high-order Boltzmann machine~\cite{sejnowski1986higher}, but employs a non-Hebbian training protocol. %In contrast with its Hebbian counterpart, which amounts to an annealed average over the dataset, the latter non-Hebbian protocol corresponds to a quenched average.
We explicitly calibrate the weights of individual configurations within the dataset and assign separate energy references to individual configurations of the machine. Thus we explicitly treat learning as contextual. %One can also make parallels with how one assigns separate references for the entropies and enthalpies of formation for distinct chemical species. In further reference to Chemistry, one may say that here we treat individual datasets as {\em mixtures} of configurations.

We have built a conjoint free energy surface that can guide, in principle, both training and retrieval. The free energy is a function of the coupling constants; it is uniquely minimized by some complete set of coupling constants whose size can be as large as $2^N$. In practice, the set of the coupling constants must be reduced in size from the said value of $2^N$, high-order couplings likelier to be removed because of their multiplicity. The resulting free energy can be thought of as a cross-section of the original surface along a sub-manifold of the original space of the coupling constants. We find that when evaluated within such sub-manifolds, the free energy has not one, but multiple minima. The latter degeneracy is connected to the degeneracy of the free energy as a function of coarse-grained degrees of freedom. This apparent connection between the free energy for learning and retrieval, respectively, can be traced to bounds on the free energy derived early on by Gibbs. We have seen that reduction in description can be thought of as a meanfield approximation. 

We have argued that in a consistent treatment, one must explicitly parameterize the energies of configurations not represented in a dataset; they cannot simply be regarded as indeterminate. Because of their huge number, the non-represented configurations must be destabilized, energy-wise, by an extensive amount relative to the represented states. Thus the non-represented configurations comprise a distinct, high temperature phase. 

%The high-temperature phase can be thought of a general pool of configurations. Retrieval of configurations that were actually present in the dataset is analogous to condensing a mixture with some non-generic composition out of a vapor that has a different composition. The aforementioned extensive energy gap corresponds to the latent heat of the transition and can be thought of as reflecting the work necessary to extract knowledge, by analogy with conventional Thermodynamics. It is conceivable that this work reflects the actual amount of energy that must be spent to perform the retrieval computationally.  Incidentally, distillation happens to be one of the most energy-consuming industrial processes~\cite{KISS2020117788}. 

Reduction in description leads to ergodicity breaking, which plays a dual role in acquisition of knowledge. Learned patterns can be thought of as a library of bound states centered at configurations from the dataset. All other patterns comprise what is essentially a high entropy continuum. Ergodicity breaking prevents escape into the continuum and, hence, is essential to discriminating correct patterns. At the same time, kinetic barriers separating distinct free energy minima in the low-temperature phase will result in kinetic bottlenecks for both learning and retrieval. The latter aspect of the ergodicity breaking is detrimental; it also seems to be an inherent feature of contextual learning. To mitigate these detrimental effects, one may have to resort to using separate generative models for distinct free energy minima. The number of minima---and hence the demand for more experts---will typically increase with the degree of reduction in description.  It is conceivable that the thermodynamic potentials considered here can be employed to rate the veracity of an individual expert, viz., by using the depth of the corresponding free energy minimum as a metric.

The above notions appear to be consistent, for instance, with the limited success machine learning-based force fields have had in predicting structures of inorganic solids. Indeed, the bonding preferences for most elements and the bond orders tend to switch among several, discretely different patterns, in a fashion similar to that shown in Fig.~\ref{mcan}. An example important in applications is the competition between the tetrahedral and octahedral bonding patterns in intermetallic alloys~\cite{ZLMicro1}. This near discreteness results from cooperative  processes \cite{GHL} that are similar to the processes causing the ergodicity breaking we  discussed here. Thus we anticipate that each distinct bonding pattern will require a separate generative model. For instance, for three atoms and two generative models per atom, there would be eight different potential outcomes for the ground state. Furthermore, the analysis in Section~\ref{falsehood} suggests that successful generative models would have to be trained on at least two {\em sets} of structures: One set corresponds to low-energy ordered structures, the other set to high-energy liquid structures. We do not, however, have access to liquid structures at present, which one might view as a fundamentally difficult aspect of the problem of predicting the structure of inorganic solids. This important problem~\cite{MaddoxEditorial1988} remains unsolved. 

The present work has focused on thermodynamic aspects of learning and retrieval. The corresponding {\em kinetics} are highly sensitive to the detailed form of the generative model and the sampling moves. A {\em generic} sampling move, during retrieval, will place the system into one of the false states essentially always, because of their multiplicity. The probability of sampling out of a false state into a true state is very very low, roughly $M_0/2^N$, corresponding to an entropic free energy barrier $N T \ln 2$. This entropic bottleneck is analogous to Levinthal's paradox of protein folding,~\cite{levinthal, WolynesNIST} according to which an unfolded protein should never find a native state, even if the latter is thermodynamically favorable. Levinthal's paradox is however resolved by noting that the landscape of the non-native states of actual proteins is not flat. Instead, its overall shape is funnel-like and minimally-frustrated notwithstanding some amount of roughness~\cite{https://doi.org/10.1002/prot.340210302, OSW, capillary}, a notion at the heart of spectrum parameterization in associated-memory Hamiltonians~\cite{doi:10.1021/jp212541y}. Likewise, it appears that machine learning will work well only if the dataset itself allows for a funneled free energy landscape or a combination of a modest number of such funnels. 

%The AMH employs interactions that are short-ranged and limited to those present in native structures.~\cite{ShojiTakada2019} This helps to reduce the critical size for nucleating the native fold and, thus, effectively guide an unfolded protein molecule toward its native state.~\cite{}

%This could be contrasted, for instance, with traditional feed-forward neural networks which effectively operate at vanishing temperatures~\cite{HeThesis} thus causing them to be irrevocably arrested in configurations, whose performance---be it good or poor---is difficult to anticipate in advance.

%One may attempt to reduce the number of bits through coarse-graining, but the resulting generative model might become too restrictive. One example of such overly restrictive generative model is provided by continuum, hydrodynamic descriptions~\cite{Gotze_MCT} of glassy liquids, which effectively assume the free energy has only one minimum and erroneously predict diverging relaxation times when the actual liquid, though slowed somewhat, still flows just fine.~\cite{LW_ARPC, HLtime} Now, 

\section*{Acknowledgments}
V.L. thanks Vitali Khvatkov, Rolf M. Olsen, and Michael Tuvim for inspiring conversations. We gratefully acknowledge the support by the NSF Grants CHE-1465125 and CHE-1956389, the Welch Foundation Grant E-1765,
and a grant from the Texas Center for Superconductivity at the
University of Houston. We gratefully acknowledge the use of the
Carya/Opuntia/Sabine Cluster and the advanced support from the Research Computing Data Core at the University of Houston acquired through NSF Award Number
ACI-1531814.

\appendix

\section{Connection with Chemistry and generalization of $\widetilde A$ for correlations in datasets.}

\label{correlations}

The notions of calibration put forth in Section~\ref{thermo} can be compared with how one counts states in Thermochemistry. In chemical contexts, one may use the semi-classical result for the density of states to normalize the partition function for a set of classical particles~\cite{LLstat}. Thereby, one counts configurations in the convention that there is one state per the volume formed by the thermal de Broglie wavelength of the particle. Since the latter wavelength is inversely proportional to the square root of the mass, one obtains that per every state of the $^{20}$Ne atom, there are $(40/20)^{3/2} \approx  2.83$ states of the $^{40}$Ar atom. But this becomes an entirely moot point in the classical regime, in which all thermodynamic properties of the system are strictly independent of mass! Consistent with this, the entropy of a classical system can be defined only up to an additive constant, and so only entropy {\em differences} are meaningful. The relevant length scale for counting states in a classical gas is the typical spacing between like-particles since a particle identity can be established only if no other particles of the same species are nearby~\cite{L_AP, LSurvey}. Thus the chemist ties standard states to densities already at the onset of the calculation; Eq.~(\ref{ZE}) has a similar purpose. %The use of standard entropies in Chemistry is a corollary of the more general notion that free energies can be defined only up to an additive constant. 

When correlations, if any, among fluctuations of the weights $x_i$ are neglected, one may determine the susceptibilities pertaining to the thermodynamic potential $\widetilde A^{(M)}$ from Eq.~(\ref{Atilde}) by Taylor-expanding the latter potential near its minimum:
\begin{align} \label{qform0}
    \widetilde A^{(M)}/T \approx & \sum^M_i \frac{(x_i-x_i^\ominus)^2}{2 x_i^\ominus} + \frac{1}{2 x'^\ominus}\left[ \sum^M_i (x_i - x_i^\ominus) \right]^2   -  \ln Z^\ominus
\end{align}
Apart from the soft constraint stemming from the normalization condition $\sum_i x_i =1$, entering as the second sum on the right, we see the susceptibilities for the individual weights $x_i$ have the characteristic form for the number fluctuations in an ideal gas, i.e. $(x_i^\ominus)^{1/2}$. Eqs.~(\ref{dAt}) and (\ref{qform0}) yield the following constitutive relations, for weak deviations from the standard model:
\begin{equation} \label{constRelIdeal}
    E_i  =  E_i^\ominus - T \ln\frac{x_i/x_i^\ominus}{(1 - \sum_i^M x_i)/x'^\ominus}.  
\end{equation}
We see that the energies $E_i$ are similar to the (negative of) the chemical potential. To avoid ambiguity, we note that unlike here, the standard state in chemical contexts usually corresponds to pure substances, $x_i^\ominus=1$. 

One can supplement the generic thermodynamic potential (\ref{Atilde}) with additional terms so as to encode correlations among the variations of the weights $x_i$, if any. At the quadratic level, one thus obtains:
\begin{equation} \label{Atc}
    \widetilde A^{(c)}/T \equiv \widetilde A/T + \frac{1}{2} \sum^M_{ij} \gamma_{ij} (x_i-x_i^\ominus)(x_j-x_j^\ominus) 
\end{equation} 
where we use the label ``$(c)$'' to distinguish the  potential modified for correlations from the basic form in Eq.~(\ref{Atilde}). Consequently, the constitutive relations become
\begin{equation} \label{Exgen}
    E_i/T = E_i^\ominus/T - \ln \frac{x_i}{x_i^\ominus} \: \frac{e^{\sum_j \gamma_{ij} (x_j - x_j^\ominus)}}{(1 - \sum_i^M x_i)/x'^\ominus} 
\end{equation}
which supersedes the simpler relation in Eq.~(\ref{constRelIdeal}). We have presented the correction as a multiplicative factor under the logarithm, in deference to the empirical laws due to Henry and Raoult, respectively~\cite{BRR, silbey2004physical}.  

Now suppose, for the sake of argument, that after we have added the correction, a quadratic expansion of the potential $\widetilde A^{(c)}$ around its minimum defines the associated probability distribution adequately. The original free energy $A$, then, must be corrected according to:
\begin{align} \label{Ac}
    A^{(c)}/T \approx & A/T - \frac{1}{2 T^2} \sum^M_{ij}  [\alpha_{ij} - \delta_{ij} x_i^\ominus +x_i^\ominus x_j^\ominus ]   (E_i-E_i^\ominus)(E_j-E_j^\ominus), \\ \approx & A^\ominus/T  - \frac{1}{2 T^2}   \sum^M_{ij}  \alpha_{ij} (E_i-E_i^\ominus)(E_j-E_j^\ominus)
\end{align} 
where the quantities
\begin{equation}
    (\alpha^{-1})_{ij} = \frac{\delta_{ij}}{x_i^\ominus} + \frac{1}{x'^\ominus}  + \gamma_{ij}.
\end{equation}
comprise, by construction, the coefficients of the combined quadratic forms from Eqs.~(\ref{qform0}) and (\ref{Atc}). $\delta_{ij}$ is the Kronecker delta function. The matrix $\alpha_{ij}$ and its inverse thus determine the correlations among fluctuations of the weights $x_i$ at the Gaussian level, Chapter 111 of \cite{LLstat}:
\begin{equation} \label{xxa}
    \la \delta (x_i-x_i^\ominus) \, \delta (x_j - x_j^\ominus) \ra = \la \delta x_i \, \delta x_j \ra = \alpha_{ij} 
\end{equation}
and of the energies, according to:
\begin{equation} \label{EEa}
    \la \delta (E_i-E_i^\ominus) \, \delta (E_j - E_j^\ominus) \ra =  (\alpha^{-1})_{ij}. 
\end{equation}
We reflected, in the first equality of Eq.~(\ref{xxa}), that the standard values $x_i^\ominus$ of the weights are fixed by construction. It is understood that the matrix $\alpha$ is positive definite. 

The coefficients $\gamma_{ij}$ in the quadratic expansion in Eq.~(\ref{Atc}) imply non-trivial correlations among fluctuations of the weights. One may imagine how such correlations can arise owing to intrinsic uncertainties in calibrating detectors. Consider, for instance, Shockley's setup of a self-guiding missile or face-recognition device~\cite{ShockleyAI}, in which the image collected by the device's camera is passed through a film containing the image of the intended target. The accuracy of the aim is assessed by measuring the intensity of the light that has passed through the film. One must set a separate intensity standard  for each individual target or even the very same target depending on the lighting conditions. A universal device, capable of processing multiple images and/or lighting conditions, would then require a floating calibration scheme for the input. Ideally the intensity standard should vary smoothly with variations in the image. Incidentally, ``elastic'' algorithms to align/match images have been discussed since the early 80's~\cite{BURR1981102, MOSHFEGHI1991271}. When alignment of two images requires deletions or insertions, the latter may be thought of as  ``lattice defects,'' by analogy with Continuum Mechanics. 

The standard values $x_i^\ominus$ are set by the calibration convention for the outputs. The choice of the source fields $E_i^\ominus$ remains flexible and, hence, can be used to implement a particular floating-calibration scheme for the inputs. The reference values $E_i^\ominus$ should be equal to each other within an error that, ideally, increases smoothly with the difference between two supposedly similar images, according to an adopted similarity criterion. (One hopes that the cost of the defects, if any, does not overwhelm the cost stemming from purely elastic distortion of defect-less portions of the image.) Thus, roughly, $(E_i^\ominus - E_j^\ominus)^2 \propto (\vec \sigma_i - \vec \sigma_j)^2$, consistently for all pairs $(i, j)$ that are neighbors in the Hamming space of configurations: $(\vec \sigma_i - \vec \sigma_j)^2 <L$ and $L$ is some judiciously chosen cutoff distance. The latter convention is analogous to the setup of a scalar field theory  defined on a discrete lattice \cite{itzykson2012quantum}; hereby the standard $E_i^\ominus$ is the field itself, while the lattice points comprise the ($N$-dimensional) Hamming space of the configurations represented in the dataset. 

The standard potentials $E_i^\ominus$ and $E_j^\ominus$ of two configurations that are further apart than the cutoff distance $L$ are still correlated, but indirectly, through chains of neighbors (in the Hamming space). The degree of correlation is problem specific: In Continuum Mechanics contexts, the average in Eq.~(\ref{EEa}) tends to a steady value in dimensions three and higher, but diverges logarithmically and linearly with the distance in two and one spatial dimensions, respectively, or in any dimensions when the shear modulus vanishes \cite{LLstat}. In any case, we conclude that owing to intrinsic uncertainties in input calibration, the coupling constants $\gamma_{ij}$ will be non-vanishing. Finally we note that the local nature of the standard state $E_i^\ominus$ is formally analogous to the locality of gauge fields in field theory~\cite{itzykson2012quantum} or, for instance, of the Berry phase in quantum mechanics~\cite{doi:10.1098/rspa.1984.0023}. Variations in such gauge fields amount to long range interactions among local degrees of freedom. Specifically in elastic continua, whether degenerate or not, such interactions are of the dipole-dipole variety~\cite{BL_6Spin, BLelast} but can be screened in the presence of fluidity~\cite{PhysRevE.104.024904}.

\section{The two-spin generative model: Calculations}

\label{spin2calculation}

There are two alternative methods to calculate the Gibbs energy, which are equivalent in an ergodic system. In one method, one evaluates the Gibbs energy starting from the Helmholtz energy (\ref{FMF3}) and then using the Legendre-transform prescription from Eqs.~(\ref{AGt}) and (\ref{hAtm}). One thus obtains
\begin{equation} \label{mh} 
\begin{split} 
    m_1 &= \tanh [( -J m_2 + h_1)/T] \\ m_2 &= \tanh [( -J m_1 + h_2)/T]
\end{split}
\end{equation}
The quantities $(- J m_\alpha)$ inside the brackets, when $m_\alpha \ne 0$, can be thought of as internal fields that emerge self-consistently. Such internal fields do emerge in non-meanfield settings as well, of course, but require cooperative effects~\cite{Goldenfeld}. 

When the Helmholtz energy $\widetilde A(m_1, m_2)$ has two minima, Eq.~(\ref{mh}) has not one, but three solutions already for a vanishing external field $h_\alpha = 0$: Two solutions correspond to the minima themselves, and one solution to the saddle point separating the minima. The two minima and the saddle point all lie within the slice $m_1 = - m_2$ from Fig.~\ref{Am1m2Figslice}. It will suffice for our purposes---and will simplify the prose a great deal---to work along the latter slice. We use $m \equiv m_1 =  - m_2$ and $h \equiv h_1 = - h_2$ as our variables.  Below the critical point, the $\widetilde A(m)$ curve exhibits two inflection points $m_\text{sp}$, whereby $(\partial^2 \widetilde A/\partial m^2)_{m_\text{sp}} = 0$. These points---conventionally called the ``spinodals''---delineate the stability limits, since the susceptibility is negative between the spinodals: $\partial m/\partial h = (\partial h/\partial m)^{-1} = (\partial \widetilde A/\partial m^2 )^{-1} < 0$.  We denote the locations of the spinodals pertaining to the positive and negative minimum as $m_\text{sp}^+$ and $m_\text{sp}^-$, respectively. When $m_\text{sp}^+ = - m_\text{sp}^- > 0$, Eq.~(\ref{mh}) will have three solutions within an $h$ interval of non-vanishing width, whose l.h.s. and r.h.s. boundaries are determined by solving Eq.~(\ref{mh}) with $m$ set at $m_\text{sp}^+$ and $m_\text{sp}^-$ respectively. This is illustrated in Fig.~\ref{mcan} with the $m_\text{MF}$ curves. Of the three solutions of Eq.~(\ref{mh}), we will consider the two stable solutions pertaining to the minima. Hereby the magnetization is subject to a bimodal distribution, the more likely mode corresponding to the deeper minimum of $\widetilde A$.

\begin{figure}[t]
\centering
\includegraphics[width=1.4\figurewidth]{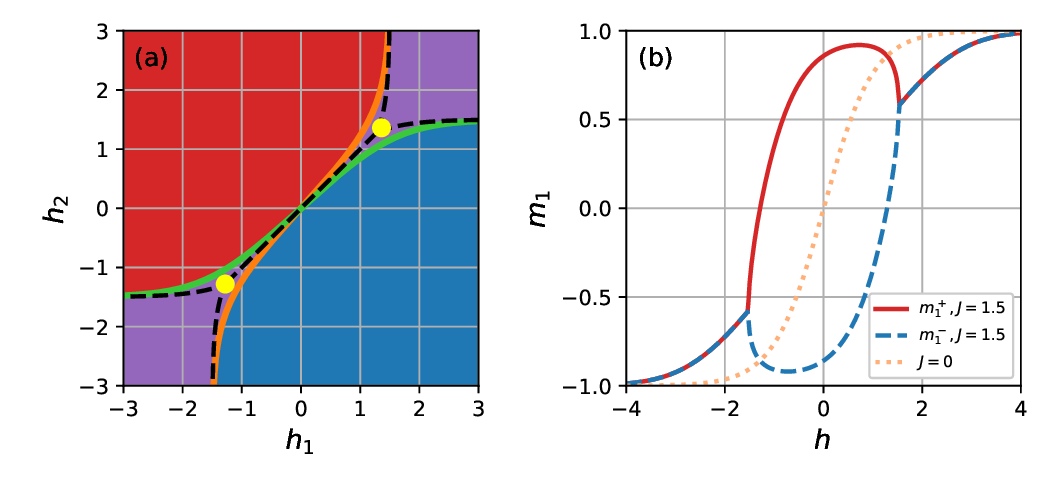}
\caption{\label{mmhhFig} (a) Signs of the exact polarizations as functions the source field for the energy function (\ref{E2}) are shown using colors. Red and blue areas: $m_1 m_2 < 0$. Purple area: $m_1 m_2 > 0$. The counterparts of the region boundaries pertaining to the likeliest meanfield magnetizations are shown using dashed black lines. The line connecting the two yellow dots corresponds to a discontinuity in the most probable value of the magnetization and, as such, is a phase boundary. (b)~The magnetization $m_1$ as a function along the phase boundary from panel (a); $h \equiv h_1 = - h_2$. The solid and dashed line to the stable and metastable minimum, respectively, of the meanfield free energy surface $\widetilde A$, at $J=1.5$, $T=T^\circ=1$.}
\end{figure}

To elucidate the nature of the meanfield constraint---which causes ergodicity breaking at sufficiently low temperatures---we first juxtapose, in Fig.~\ref{mmhhFig}(a), the signs of the magnetizations $m_\alpha$ as functions of the source fields $h_\alpha$ for the exact and meanfield solution, respectively, of the generative model (\ref{E2}). For the exact solution, we color-code the regions of positive and negative $m_\alpha$ with red and blue, respectively. The purple areas, then, show where the exact values of the two typical polarizations $m_1$ and $m_2$, respectively, have the same sign. For the meanfield solution, possible values of the magnetizations are determined by the positions, in the $(m_1, m_2)$ plane, of the minima of the tilted free energy surface $\widetilde A(m_1, m_2) - m_1 h_1 - m_2 h_2$, where the fields $h_\alpha$ are treated as constants, c.f. the discussion following Eq.~(\ref{hAtm}). We show, in Fig.~\ref{mmhhFig}(a), the signs for the likelier magnetization pattern, viz., the one corresponding to the deeper minimum of the tilted surface. The corresponding boundaries are shown using dashed lines; we see they lie rather close to their exact counterparts. Unlike the signs, the {\em magnitudes} of the exact and meanfield magnetizations, respectively, show a qualitatively different behavior except when the source fields are large enough for the (tilted) meanfield free energy surface to exhibit just one minimum. (We reiterate that the exact free energy always has just one minimum.)  Consequently, the likeliest values of meanfield magnetizations experience a discontinuity when the two competing minima of the free energy surface are exactly degenerate. Conditions for such degeneracy are met along a substantial segment of the $h_2 = h_1$ line shown in Fig.~\ref{mmhhFig}(a) as the straight dashed line connecting the two yellow dots. The latter segment thus represents a {\em phase boundary}. An elementary calculation shows the ends of the latter phase boundary are located at $(\pm h_c, \pm h_c)$, where $h_c = J m_0 + T \text{atanh} (m_0)$ and $m_0 \equiv (1-T/J)^{1/2}$, while the meanfield $m_\alpha = 0$ lines are given by functions $h_\alpha/T = \text{atanh} (h_\beta/J)$, $|h_\alpha| \ge h_c$.

The distinct difference of the likeliest magnetizations on the two opposite sides of the phase boundary---the $m_1$ component shown in Fig.~\ref{mmhhFig}(b)---is an instance of hysteresis, a classic signature of broken ergodicity. The discontinuity across the phase boundary implies that a substantial region of phase space around the origin $\vec m = 0$ is strictly avoided for sufficiently large values of the coupling $J$, as a result of the meanfield constraint. %Consequently, the portion of the meanfield free energy surface around the origin does not move down even as the coupling is increased. This causes a barrier to appear when the coupling $J$ constant becomes large enough for the energy term, $J m_1 m_2$, to begin overtaking the entropy term for $| m_i | \simeq 1$. 

An alternative way to compute the Gibbs energy is directly through the equilibrium partition function with and added source field, Eq.~(\ref{Gcg}). The results of this calculation for the exact and meanfield case, respectively, are shown in the main text. Here we only verify that the setup in Eq.~(\ref{Gcg}) yields the meanfield description (\ref{FMF3}) in the $N_r \to \infty$ limit.  First we note a Hubbard-Stratonovich formula, Eq.~(27.55) from Ref.~\cite{schulman2012techniques}, that can be used to uncouple the two factors in the product in Eq.~(\ref{Exz}):
\begin{equation}
    e^{\xi^* \zeta} = \int \frac{d^2 \eta}{\pi} e^{-\eta^* \eta + \xi^* \eta + \eta^* \zeta}.
\end{equation}
Here $d^2 \eta = d(\text{Re } \eta) d(\text{Im } \eta)$ corresponds to integration in the complex plane and the integration variables $\eta$ and $\eta^*$ are treated as independent. One may now straightforwardly sum over the spin states to obtain:
\begin{align} \label{Gcg2}
    \widetilde G_\text{eq} & =  - T \ln \left( \int \frac{d^2 \eta}{\pi} \exp \left\{N_r  \left[ - \frac{J \eta^* \eta}{T} + 2 \ln 2 \right. \right. \right. \nonumber \\ & + \left. \left. \left. \ln \cosh \left( \frac{J \eta + h_1}{T} \right) +    \ln  \cosh \left( \frac{J \eta^* - h_2}{T} \right)  \right] \right\} \right). 
\end{align}
The $N_r \to \infty$ asymptotics can be readily evaluated by the saddle-point integration. The locations of the stationary points $(\eta_0, (\eta^*)_0)$ are solutions of the following system of equations:
\begin{equation} \label{etaeta}
    \begin{split}
         (\eta^*)_0   &= \tanh \{ [ - J (-\eta_0) + h_1]/T \} \\ -\eta_0 &= \tanh \{ [- J (\eta^*)_0  + h_2]/T \}.
    \end{split}
\end{equation}
To avoid confusion we note that generally $(\eta^*)_0  \ne (\eta_0)^* $. To each saddle point there corresponds a Gibbs energy whose value, in the leading order in $N_r$, equals:
\begin{equation} \label{GhNr}
\begin{split}
    \frac{\widetilde G}{N_r}  =  \frac{J}{T} (\eta^*)_0 \eta_0 - \ln 2 \cosh \left( \frac{J\eta_0  + h_1}{T} \right)  - \ln  2 \cosh \left( \frac{J (\eta^*)_0  - h_2}{T} \right).     
\end{split}
\end{equation}
Comparing the derivatives of Eq.~(\ref{Gcg}) and Eq.~(\ref{Gcg2}), respectively, with respect to the source fields allows one to identify $(\eta^*)_0$ as $m_1$ and $(-\eta_0)$ as $m_2$, in the $N_r \to \infty$ limit. Thus Eq.~(\ref{etaeta}) is identical to Eq.~(\ref{mh}), while the branches $\widetilde G^+$ and $\widetilde G^-$, respectively, of the restricted reduced Gibbs energy must be identified with the two stationary points of the integrand in Eq.~(\ref{Gcg2}). Consistent with this identification, the formula $\ln 2 \cosh(x) = -\{(1+y) \ln [(1+y)/2] + (1-y) \ln [(1-y)/2] \}/2 + xy$, where $y \equiv \tanh x$, allows one to see that the Gibbs energy per replica $\widetilde G/N_r$ from Eq.~(\ref{GhNr}), when computed for an individual branch, equals precisely the restricted Gibbs energy $\widetilde G$ from Eq.~(\ref{AGt}). (The equilibrium value $\widetilde G_\text{eq}$, in the $N_r \to \infty$ limit, is equal to the lowest of the set of $\widetilde G$'s from Eq.~(\ref{GhNr}) as computed for the full set of the stationary points.)

\section{Ergodicity breaking is transient. It is restored via rare, cooperative processes}

\label{transient}

\begin{figure}[t]
\centering
\includegraphics[width= 0.85\columnwidth]{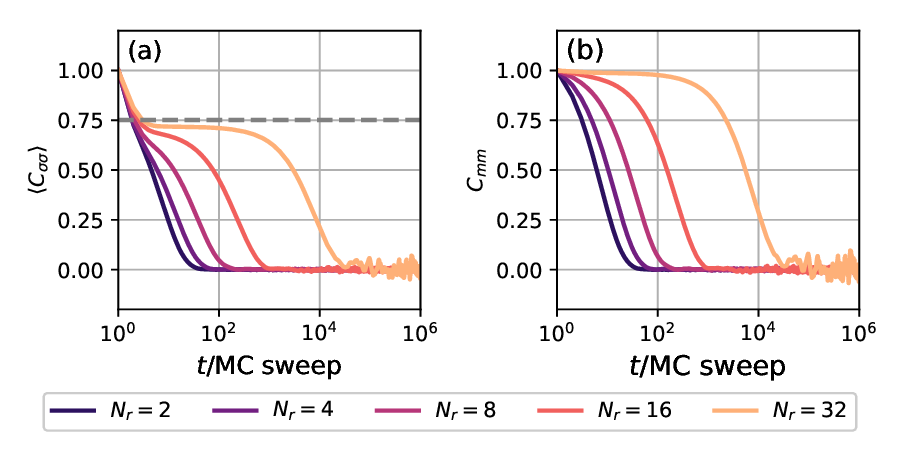}
\caption{\label{corrNr} Relaxation profiles for equilibrium Monte Carlo simulations. (a) Single spin autocorrelation function $C_{\sigma \sigma} \equiv \la \xi_s (t+t_0) \xi_s (t_0) \ra$ averaged over individual spins and over $t_0$. The horizontal dashed line indicates the location squared of an individual minimum of $\widetilde A(m)$, which is the limiting height of the plateau when the escape time from an individual minimum diverges. (b) Multi-spin autocorrelation function $C_{mm} \equiv \la \sum_s \xi_s(t+t_0) \sum_q \xi_q (t_0) \ra/N_r^2$. $J=1.5$}
\end{figure}

In a finite system, the ergodicity is eventually restored. To this end, we show in Fig.~\ref{corrNr} autocorrelation functions for two select observables; the time is measured in steps of a Monte Carlo simulation.  In panel (a), we display the single spin-spin autocorrelation function $C_{\sigma \sigma} (t) \equiv \la \xi_s (t+t_0) \xi_s(t_0) \ra$, where the averaging is over the location $t_0$ of the sampling window and over the spins. Past a certain, modest value of the system size, we clearly observe two distinct, time-separated processes: The short-time process corresponds to the vibrational relaxation within an individual free energy minimum; call the corresponding relaxation time $t_\text{vib}$. For $t > t_\text{vib}$, the correlation function temporarily settles at a plateau value $\la \xi_s (t+t_0) \xi_s (t_0) \ra \lesssim \la \xi \ra_\text{vib}^2$, where the average $\la \xi \ra_\text{vib} \ne 0$ pertains to an individual minimum and $t$ is less than the typical escape time $t_c$ from a minimum. ($\la \xi_s (t+t_0) \xi_s (t_0) \ra \to \la \xi \ra_\text{vib}^2$ for $t_c \to \infty$ and long times $t$, $t_\text{vib} < t < t_c$.) In the long term, viz. $t > t_c$, the system is however able to transition between the two minima; as a result the average magnetization eventually attains its equilibrium value of zero $\la \xi \ra = 0$. This causes the correlation function to decay to zero, too. The inter-minimum dynamics is sometimes called configurational relaxation~\cite{HLtime, LW_ARPC, L_AP}. The appearance of time-scale separation between the vibrational and configurational relaxation, respectively, implies ergodicity is broken, even if transiently. The length of the plateau then reflects the temporal extent of the ergodicity breaking. The latter extent scales exponentially with the parameters of the problem and can become very long already for modestly sized systems and/or following small temperature variations. We confirm the cooperative nature of ergodicity-restoring transitions in panel (b) of Fig.~\ref{corrNr}, where we display the autocorrelation function $$C_{mm} \equiv \la \sum_s \xi_s(t+t_0) \sum_q \xi_q (t_0) \ra/N_r^2.$$ The  vibrational component of the relaxation is largely averaged out already at the onset implying the overall relaxation is largely due to inter-minimum transitions.     

\section{Ergodicity breaking causes the parameter manifold to fractionalize into a set of fragments that are compact}
\label{shape}

Below the critical point, the minima of the surface $\widetilde A(m_1, m_2)$ from Eq.~(\ref{FMF3}) are strictly degenerate when $J_1^\ominus$ and $J_2^\ominus$ vanish. The latter degeneracy is dictated by the invariance of the product $\sigma_1 \sigma_2$ with respect to flipping the two spins at the same time. Yet the latter symmetry has another consequence, viz., that the average magnetizations $m_\alpha$ must all vanish at all temperatures. At the same time, the operation $(\sigma_1, \sigma_2) \to (-\sigma_1, -\sigma_2)$ is intrinsically discrete. Thus the symmetry breaking signalled by the emergence of the two minima in $\widetilde A(m_1, m_2)$---each of which corresponds to {\em non-vanishing} magnetizations $m_\alpha \ne 0$---must be {\em also} of the discrete kind. Consequently, no Goldstone modes~\cite{PhysRev.127.965} appear as a result of the symmetry breaking; the newly emerged minima of the free energy must be separated by a barrier. The barrier can be made small near criticality, if any, but criticality is ordinarily observed within a manifold of vanishing volume in the phase space and, thus, is rare. 

One may generalize the above discussion to higher order interactions by replacing the object $\sigma_2$, in model (\ref{E2}), by the object $\sigma_2 \sigma_3$, while keeping the overall coupling constant positive. The resulting standard generative model:
\begin{equation} \label{E123}
    E^\ominus = J \sigma_1 \sigma_2 \sigma_3
\end{equation}
now corresponds to the logic operation XOR:
\begin{equation}
    \begin{tabular}{c|cc} 
     & $+1$ & $-1$  \\ \hline 
         $+1$ & $-1$ & $+1$  \\ 
         $-1$ & $+1$ & $-1$ 
    \end{tabular}
\end{equation}
where the labels of the rows and columns correspond to the respective states of any two spins while the entries in the body of the table correspond to the states of the remaining spin. 

The product $\sigma_2 \sigma_3$ %forms a representation of the group Z$_2 \times$Z$_2$, as already mentioned, but also 
can be presented as a direct sum Z$_2 \, \oplus\,$Z$_2$. This amounts to the three-body energy function (\ref{E123}) being a sum of two equivalent, non-interacting replicas of the {\em two}-body energy function from Eq.~(\ref{E2}). Thus one may consider a trial description with two couplings $J_{1}$ and $J_{23}$:
\begin{equation} \label{MFansatz123}
    E = - J_1 \, \sigma_1 - J_{23} \, \sigma_2 \, \sigma_3,
\end{equation}
where the compound object $\sigma_2 \sigma_3$ can be treated as a single spin while the overall free energy is lowered by $T \ln 2$, relative to a two-spin system proper. Fluctuations of spins 2 and 3 are correlated with each other, but not so with fluctuations of spin 1:
\begin{equation} \label{MFconstr3}
    \overline{\sigma_1 \sigma_2 \sigma_3} = \overline\sigma_1 \: \overline{ \sigma_2 \sigma_3}.
\end{equation}
The preceding discussion can be repeated to see that the factorization scheme (\ref{MFconstr3}) will result in a pair of identical, doubly-degenerate free energy surfaces. Interactions of order four and higher can be treated analogously. In effect, correlations of the respective rank will be approximated as products of lower-rank correlation functions, a common meanfield approximation~\cite{Zubarev}. 

When present, Goldstone modes imply the free energy minima, below the symmetry breaking, have a vanishing curvature along one or more directions in the order-parameter space. We see such non-compact free energy minima would be untypical for binary datasets thus indicating the ergodicity breaking is of the harshest type possible. 

For many problems, considering digitized datasets as binary is arguably gratuitous, especially when the underlying problem is continuous. In such cases, Goldstone modes would be absent nonetheless. This lack of Goldstone modes can be viewed, rather generally, as a consequence of the symmetry of the contextual ensemble (\ref{ZE}) with respect to the gauge transformation $E_i^\ominus/T \to E_i^\ominus/T + \delta_i$, $\ln Z_i^\ominus \to \ln Z_i^\ominus - \delta_i$. When gauged symmetries are broken, candidate Goldstone excitations do become gapped \cite{AndersonBasicNotions, itzykson2012quantum}. Now, a distinct variety of low-frequency modes can arise in models with translationally invariant, short-range forces when more than one phase coexist and are in near equilibrium with each other. Hereby, the system is broken up into regions each occupied by an individual phase, a phenomenon called ``spinodal decomposition''~\cite{Goldenfeld, Bray}. Interfaces separating the latter regions can often deform and move about with relative ease. We view such situations as coincidental because conditions for phase equilibrium could be fulfilled only within a manifold of vanishing volume, in the phase space.

\section{Patterns vs. generic configurations: A thermodynamic view}
\label{FTcoexist}

The restricted entropy $S_\text{F}(E_\text{F})$, corresponding to Eqs.~(\ref{ZF}), (\ref{Ecan}) and (\ref{Scan}), can be parameterized to be a strictly convex-up function by construction. Assume for now that the entropy $S_\text{T}(E_\text{T})$ of the true states---computed using Eqs.~(\ref{ZT}), (\ref{Ecan}) and (\ref{Scan})---is {\em also} a strictly convex-up function. (We use distinct variables, $E_\text{T}$ and $E_\text{F}$, for the energies of the true and false states, because their respective spectra generally overlap.) When the energy gap is at its lowest allowed value---corresponding to the equality in Eq.~(\ref{Egapmin})---the two phases are in mutual equilibrium. Indeed, $(\partial S_\text{T}/\partial E_\text{T}) = (\partial S_\text{F}/\partial E_\text{F}) = 1/T^\circ$ when evaluated at $E_\text{T}(T^\circ)$ and $E_\text{F}(T^\circ)$, respectively. Note that Eq.~(\ref{Egapmin}), at equality, expresses the familiar double tangent construction for phase equilibrium in the microcanonical ensemble \cite{Lcompeting, LSurvey}. 

\begin{figure}[t]
    \centering
    \includegraphics[width=1.2\figurewidth]{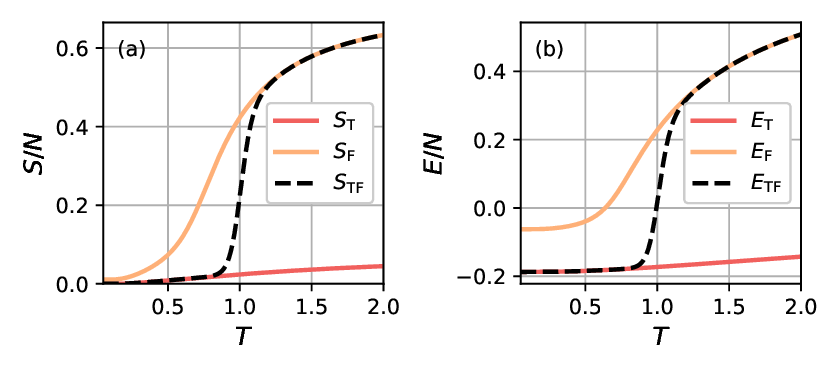}
    \caption{\label{SET} Canonical entropy and energy as functions of temperature for the discrete energy distribution $\Omega(E)$ shown as dots in Fig.~\ref{blocks}. The restricted, single-phase averages over the lower-energy block are labeled with ``T'', over the higher-energy block is labeled with ``F''. The full, equilibrium quantities are labeled with ``TF''.  $N=64$.}
\end{figure}

The two circles in Fig.~\ref{blocks} correspond to $T=T^\circ$. The mutual arrangements of the true and false states in Fig.~\ref{blocks} represents the borderline case: Moving the false states by any amount toward lower energies---while keeping $\Omega_\text{T}(E_\text{T})$ fixed--- would make them more stable than the true states. Conversely, one {\em is} allowed to use a gap that is greater than the one we used in Fig.~\ref{blocks}. For this reason, $T_0 \ge T^\circ$. Now suppose that one has settled on a specific spectrum for the false states that satisfies the constraint (\ref{Egapmin}). The two phases will be at equilibrium at a temperature $T_0$ from Eq.~(\ref{T0}). Because of the discrete change of the entropy at the transition, $\Delta S \equiv S_\text{F}(T_0) - S_\text{T}(T_0) > 0$, the transition is discontinuous and, furthermore, exhibits a {\em latent heat} equal to $T_0 \Delta S$.

\begin{figure}[t]
    \centering
    \includegraphics[width=1.4\figurewidth]{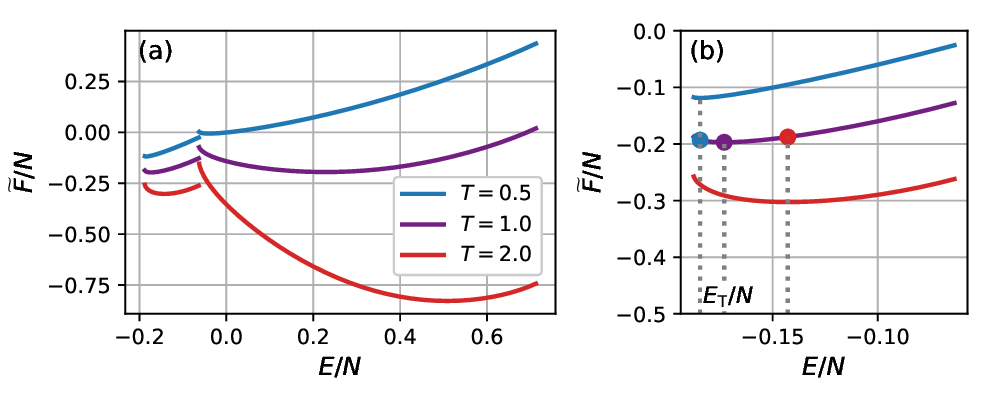}
    \caption{\label{AtEFig} (a) Thermodynamic potentials $\widetilde F(E) \equiv E - T S(E)$, where $T$ is an externally fixed temperature, for the true and false states. The restricted entropies are the same as in Fig.~\ref{blocks}. Panel (b) provides an expanded view of the low-$T$ phase to illustrate that in order to sample ``true'' states above $E_\text{T}(T_0)$ one must raise the temperature above $T_0$ and vice versa for the states with $E < E_\text{T}(T_0)$. The main graph shows that at $T > T_0$, the ``true'' states are however only metastable.} 
\end{figure}

Criterion (\ref{TT0}) can be lucidly visualized by plotting, on the same graph, the thermodynamic potential $\widetilde F(E) \equiv E - T S(E)$ for each of the individual phases, where the temperature $T$ is now regarded as a fixed, externally imposed parameter. For a single-phase system the potential $\widetilde F(E)$, as a function of $E$, is uniquely minimized at $E$ such that $\partial \widetilde F/\partial E = 1 - T \partial S/\partial E = 0$; hereby the internal temperature $(\partial S/\partial E)^{-1}$ becomes equal to $T$ \cite{LSurvey}. The depth of the minimum is equal to the equilibrium Helmholtz energy $A(T) = E(T)-TS(T)$. During phase coexistence, there is a separate $\widetilde F(E)$ minimum for each phase; the deepest minimum corresponds to the stable phase. We observe directly in Fig.~\ref{AtEFig}(a) that at $T<T_0$ the true states are more stable than the false states and vice versa at $T>T_0$.

The {\em equilibrium} energy $E_\text{eq}(T)$ and entropy $S_\text{eq}(T)$ correspond to the full partition function $Z_\text{TF}$ from Eq.~(\ref{Zeq}); they are shown in Figs.~\ref{blocks} and \ref{SET} with the dashed line. We observe that $E_\text{eq}(T)$ and $S_\text{eq}(T)$  each undergo an abrupt variation within a narrow temperature interval already at the modest value $N=64$ of the system size. It is straightforward to show that in the thermodynamic limit, $N \to \infty$, the equilibrium energy and entropy develop a strict discontinuity at $T=T_0$, consistent with Fig.~\ref{SET}, while the parametric equilibrium entropy $S_\text{eq}(E_\text{eq})$ asymptotically tends, within the gap, to the common tangent to the respective entropies of the pure phases, consistent with Fig.~\ref{blocks}. In any event, the equilibrium $S_\text{eq}(E_\text{eq})$---which is a convex-up envelope of the two restricted entropies, respectively---does not pertain to either one of the individual phases when $E \in [E_\text{T}(T_0), E_\text{F}(T_0)]$.  In summary, the entropy and energy alike can not be regarded as one-valued state functions during phase coexistence, analogously to the discussion of ergodicity breaking in Section~\ref{reduced}.

% The \nocite command causes all entries in a bibliography to be printed out
% whether or not they are actually referenced in the text. This is appropriate
% for the sample file to show the different styles of references, but authors
% most likely will not want to use it.
% \nocite{*}

\bibliographystyle{apacite}
%\bibliography{ref,lowT}% Produces the bibliography via BibTeX.

\end{document}